# Automating IRAC Analysis in Malaysian Contract Law using a Semi-Structured Knowledge Base

**Xiaoxi Kang**[1] · **Lizhen Qu**[2] · **Lay-Ki Soon**[1] · **Zhuang Li**[4] · **Adnan Trakic**[3]

**Abstract**
The effectiveness of Large Language Models (LLMs) in legal reasoning is often limited due to the unique legal terminologies and the necessity for highly specialized knowledge. These limitations highlight the need for high-quality data tailored for complex legal reasoning tasks. This paper introduces LegalSemi, a benchmark specifically curated for legal scenario analysis. LegalSemi comprises 54 legal scenarios, each rigorously annotated by legal experts, based on the comprehensive IRAC (Issue, Rule, Application, Conclusion) framework from Malaysian Contract Law. In addition, LegalSemi is accompanied by a structured knowledge base (SKE). A series of experiments were conducted to assess the usefulness of LegalSemi for IRAC analysis. The experimental results demonstrate the effectiveness of incorporating the SKE for issue identification, rule retrieval, application and conclusion generation using four different LLMs.

## 1 Introduction

Ensuring access to justice remains a pervasive challenge across societies worldwide. Two-thirds of people in the United States experienced at least one legal issue in the past four years, with less than half of those problems completely resolved.[1] In India, more than 10,490 legal cases in the Supreme Court of India have been pending for more than a decade (Madhana and Subhashree 2022). These backlogs are often caused by the complexity in legal practice, as well as the scarcity of legal professionals. IRAC framework (Metzler 2002), stands for issue, rule, application, and

[1] https://iaals.du.edu/publications/justice-needs-and-satisfaction-united-states-america

Extended author information available on the last page of the article conclusion, is the problem solving framework widely used by legal professionals to determine the underlying legal issues, followed by extracting and transforming facts in a legal scenario for legal reasoning, which eventually leads to a legal conclusion.

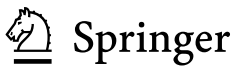

Artificial Intelligence (AI) models, in particular Large Language Models (LLMs), demonstrate great potential to improve access to justice (Krasadakis et al. 2024). However, it remains a challenge for LLMs to perform IRAC analysis on legal scenarios accurately. A recent study (Kang et al. 2023; Harasta et al. 2024) identifies two key problems in analysing legal scenarios. First, ChatGPT draws wrong conclusions on approximately 50% of the legal scenarios on average. Even if the conclusions are correct, there are mistakes in the intermediate reasoning steps. Secondly, ChatGPT is not able to cite correct legal rules when analysing the majority of the legal scenarios. In the real world, it is crucial for legal professionals to understand every single reasoning step that leads to the final conclusion. In addition, our empirical study finds that LLMs struggle to cope with the language gaps between legalese and everyday language. We conjecture that LLMs still cannot fully comprehend the underlying legal knowledge and perform complex legal reasoning accurately.

Recent advances show that it is possible to mitigate the hallucination problem of LLMs by leveraging Semi-structured Knowledge Base (SKEs) (Pan et al. 2024). SKEs can enhance LLMs in terms of interpretability and faithfulness by providing external knowledge (Kim et al. 2024). The SKEs enables easy maintenance of the legal knowledge, it is also easy to keep it up-to-date, in accordance with the revisions of legislation. Unfortunately, existing IRAC datasets lack SKEs for legal knowledge.

To address the problems above, we curate LegalSemi,[2] a dataset comprising legal scenarios pertaining to *Malaysian Contract Law*, accompanied by rich structured IRAC analysis, carried out by top law students, as illustrated in Fig. 1. Compared to Kang et al. (2023), their dataset is not only extended by doubling the legal scenarios in Malaysian Contract Law, new annotation types were also introduced to all 54 scenarios, including legal concepts and court cases. The inter-annotator agreement across all scenarios exceeds 0.8.

To support reasoning with structured legal knowledge, semantic information from a law textbook and legislation were extracted automatically to build a Semi-structured Knowledge Base (SKE). In the SKE, a node represents either a legal concept, a court case, a legal rule, the interpretation of a legal rule or a concept in lay language, or relevant meta information, while an edge between two nodes denotes their relation. The rigorous structure in the textbook and the legislation facilitates rule-based extraction of these semantic relations. The SKE is considered semi-structured because a node can be represented as either a symbol or a legal text that precisely articulates the relevant rules, cases, and concepts etc.. In contrast, SKEs typically represent all nodes and edges as symbols, a practice that, as noted in Kang et al. (2023), is challenging for LLMs to comprehend and process accurately.

The SKE facilitates issue identification, rule retrieval, application and conclusion generation by supporting reasoning over *linked legal knowledge*. Each task involves a reasoning process that begins by identifying legal concepts within given scenarios, followed by multi-hop reasoning that traverses specific rules, law sections, and rel-evant case law linked to these concepts. For example, when analyzing a scenario where 'Vanessa advertised a limited edition vinyl for $500,' the system identifies the legal concept 'invitation to treat' within the SKE, which is linked to relevant case law such as Eckhardt Marine GMBH v Sheriff [2001] 4 MLJ 4 (CA). Given the issues in that case, an LLM generates the legal issue: 'Whether Vanessa's advertisement constitutes an invitation to treat or an offer?', which is further connected to the established rule in the SKE that advertisements are generally considered invitations to treat, not binding offers (as established in Carlill v Carbolic Smoke Ball Co). Through these semantic connections, the system performs reasoning by navigating through appropriate rules, concepts, and interpretations in the SKE. It demonstrates how the legal knowledge structure in the SKE, derived from the well-defined organization of legal textbooks and legislation, enables effective IRAC analysis.

[2] https://anonymous.4open.science/r/legalsemi-3DFB/

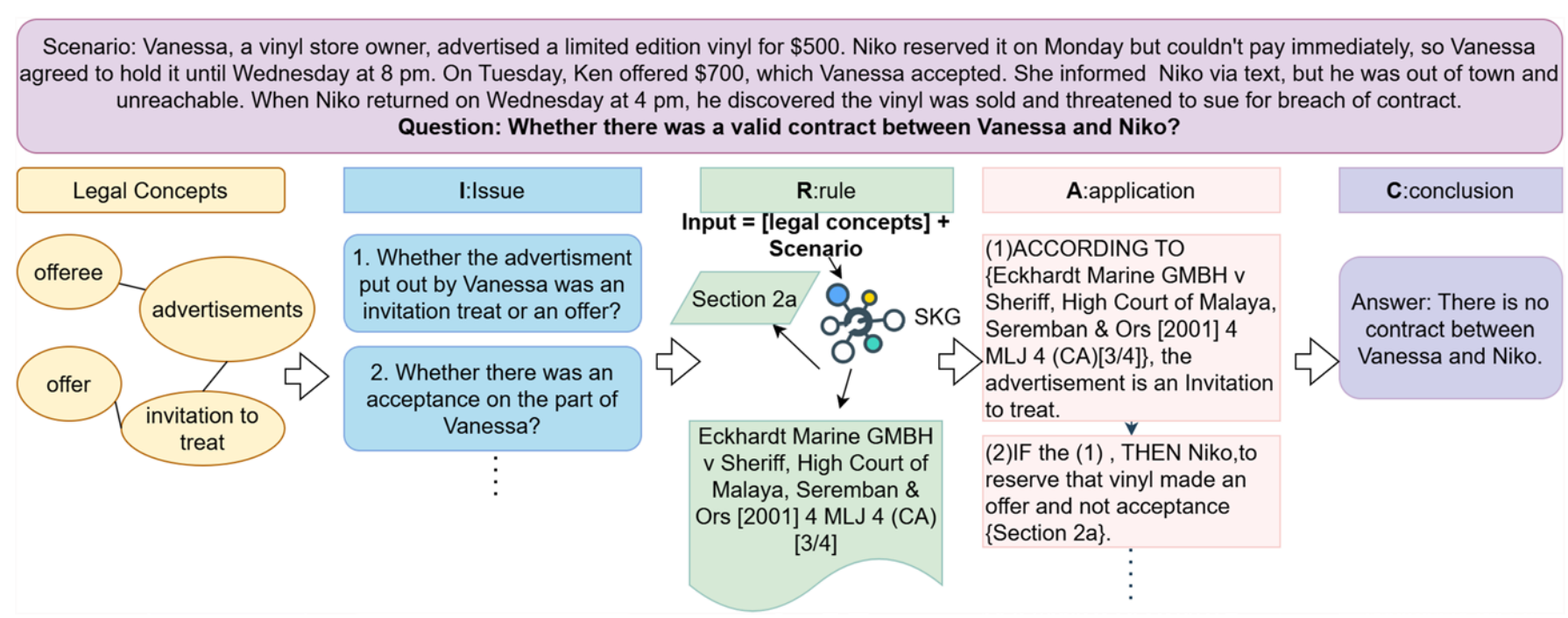

**Fig. 1** An example of a legal scenario pertinent to Malaysian Contract Law with annotations for IRAC analysis. The new types of annotations are legal concepts, court cases, and links to SKEs

We conducted extensive experiments to evaluate LLM performance on each IRAC task, finding that the SKE is particularly effective for issue identification and rule retrieval.

- Following Kang et al. (2023), we evaluate the capability of LLMs on identifying a set of issues pertinent to a legal question. The key difference to the prior work is incorporating legal concepts from the SKE, which improves the quality of issue generation by over 21.4% across all evaluated LLMs. GPT-3.5 Turbo demonstrated the most substantial improvement, with a 28% performance increase when leveraging these legal concepts for issue identification.
- By incorporating the SKE into the rule retrieval process, the results were improved significantly, increasing the performance from 6.32% to 12.93%. Comparing different models for the retrieval task across both statutes and cases, Smooth TF-IDF performed the best by using indices constructed with enriched representations. Those representations include interpretations and examples from the SKE to mitigate the linguistic disparity between legalese and lay language.

## 2 Related work

### 2.1 Legal reasoning benchmarks

Legal reasoning presents unique challenges for Natural Language Processing (NLP) systems, requiring a combination of factual understanding, rule identification, and logical application. Several benchmarks introduce these challenges. SARA (Holzenberger et al. 2020) specifically targets statutory reasoning assessment, requiring systems to interpret legal statutes and apply them to factual scenarios. CaseHOLD (Zheng et al. 2021) focuses on identifying holding statements in US case law, which represent core legal principles and reasoning outcomes. ContrActQA (Aejas et al. 2024) evaluates contract understanding through question answering, requiring systems to reason about contractual terms and implications.

For regulatory and policy understanding, PrivacyQA (Ravichander et al. 2019) assesses reasoning capabilities over privacy policies, while CUAD (Hendrycks et al. 2021) tests systems' ability to extract and reason about contractual clauses. The LeX-Files benchmark (Chalkidis et al. 2023) approaches legal reasoning from an evidence perspective, requiring systems to retrieve relevant evidence and verify legal claims.

In the broader domain of legal reasoning across jurisdictions, the COLIEE competition (Rabelo et al. 2022) has been instrumental in advancing legal entailment and case retrieval through its annual shared tasks. The recent LegalBench (Guha et al. 2023) provides a comprehensive evaluation framework with over 100 tasks specifically designed to assess legal reasoning capabilities across multiple dimensions, including statutory interpretation, contract analysis, and case outcome prediction.

Despite these advances, a critical gap remains in current legal NLP research. While the aforementioned datasets cater to different legal systems and tasks, none explicitly annotate complete reasoning steps or adopt a comprehensive IRAC (Issue, Rule, Application, Conclusion) framework. This limitation is particularly significant as reasoning capabilities are fundamental to solving legal problems effectively. Our work addresses this gap by developing a framework that explicitly models the legal reasoning process through structured knowledge bases and explicit reasoning steps.

### 2.2 Information extraction for legal NLP

Rule retrieval is a critical task in legal research that requires sophisticated information extraction and document segmentation techniques. The ability to accurately identify and extract key legal elements from complex documents forms the foundation for effective legal reasoning systems.

Kwak (n.d.a) conducted comprehensive work evaluating GPT-4's capabilities in extracting information from legal wills. Their research assessed how effectively large language models could identify and extract critical elements such as beneficiaries, executors, and specific bequests. The findings revealed both promising capabilities and significant limitations of GPT-4 in handling complex legal documentation, particularly when identifying relationships between entities and understanding hierarchical document structure.

Building on this foundation, Kwak (n.d.b) introduced an innovative "Classify First, and Then Extract" technique for legal information extraction. This prompt chaining methodology significantly improves extraction accuracy by first classifying document sections before attempting detailed information extraction, thereby providing essential context for the extraction task. This approach has demonstrated particular effectiveness for complex legal documents where contextual understanding and document structure critically impact information extraction performance.

For contract analysis, Hendrycks et al. (2021) developed a dataset for extracting information from 150 legal contracts, while Spyretta Leivaditi and Rossi (2020) employed 179 lease agreements with similar annotation schemes. Though valuable contributions, these datasets remain limited in scope compared to more recent extraction-focused approaches. In the domain of statute retrieval, Parashar (Kim et al. 2021) investigated retrieval techniques for Central Acts in the Indian Parliament, demonstrating that LemurTF-IDF substantially improved performance for statute retrieval tasks.

Despite the proliferation of legal datasets, most concentrate primarily on document labeling and classification problems rather than deeper information extraction. These datasets typically lack the features and annotations necessary to support sophisticated legal reasoning tasks. Existing works closest to our research focus predominantly on statute retrieval, which presents limitations for legal systems based on case law, where judicial decisions carry equal importance to statutory provisions. Our work addresses this gap by developing more comprehensive information extraction techniques applicable to diverse legal document types.

### 2.3 Legal reasoning approaches

Legal reasoning brings unique challenges to the NLP system. This requires understanding application in the legal process. Especially with the involvement of the LLM, now the "Reasonable" system is more critical in logic.

Legal Prompting (Yu et al. 2022) and Kang et al. (2023) is a technique to guide the LLM to think like lawyers. This approach structures prompts to follow legal reasoning patterns, breaking down complex legal questions into manageable components that align with how legal professionals approach problems. Their work demonstrated that proper prompting strategies could significantly improve the performance of language models on legal reasoning tasks without requiring domain-specific fine-tuning.

Jiang and Yang (2023) introduced syllogism prompting for the legal judgement prediction. This approach adapts the classical syllogistic reasoning structure to the legal domain, instructing language models to identify major premises (legal rules), minor premises (facts of the case), and reach conclusions through deductive reasoning. Their experiments showed that syllogism-based prompting significantly improved accuracy in predicting legal judgments compared to standard prompting techniques.

Deng et al. (2023) further expanded the application of syllogistic reasoning for legal judgment analysis. Their work established a comprehensive framework for breaking down complex legal cases into a series of interconnected syllogisms, enabling more transparent and traceable reasoning processes. This approach not only improved prediction accuracy but also enhanced the explainability of model outputs, a critical consideration in legal applications.

Prakken (2024) provided a critical analysis of methodologies for evaluating the legal reasoning capabilities of generative AI. Their work emphasized the importance of distinguishing between different types of legal reasoning (statutory interpretation, case-based reasoning, teleological reasoning) and proposed evaluation frameworks that account for the nuanced nature of legal argumentation. This research highlighted that evaluation methods must reflect the complexity and diversity of legal reasoning tasks.

Nguyen et al. (2024) investigated how large language models' reasoning capabilities could be leveraged to infer implicit concepts in legal information retrieval. Their approach utilized guided reasoning techniques to help models identify unstated but legally relevant concepts in queries and documents, significantly improving retrieval performance for legal information systems. This work demonstrated that explicit reasoning steps could bridge gaps between legal queries and relevant documents.

Yuan et al. (2024) examined whether large language models could grasp legal theories through a multiagent collaboration framework. Their approach simulated legal deliberation by having multiple language model instances assume different roles (judge, prosecutor, defense attorney) and engage in structured dialogue. This collaborative reasoning approach enabled more comprehensive legal analysis by incorporating multiple perspectives and checking for logical inconsistencies across different legal viewpoints.

These studies collectively demonstrate the evolution of approaches to enhance legal reasoning in language models, progressing from basic prompting techniques to sophisticated syllogistic frameworks and multi-agent deliberation systems. While significant advances have been made, challenges in handling complex legal reasoning that requires deep domain knowledge, understanding legal nuance, and producing legally sound arguments remain challenging, particularly to present the outcomes in a form where the legal practitioners would find it useful and trustworthy.

### 2.4 Structured knowledge graph

Knowledge graphs (KGs) have emerged as powerful tools for enhancing language models with structured external knowledge. Formally, a knowledge graph can be defined as a structured representation of facts, consisting of entities as nodes and relationships as edges, typically expressed as triples in the form of (subject, predicate, object) (Hogan et al. 2021). SKILL (Moiseev et al. 2022) demonstrated significant improvements by integrating pre-trained models with the Wikidata KG (Vrandecic and Krotzsch 2014), outperforming T5 (Raffel et al 2020) baselines on multiple question-answering benchmarks including FreebaseQA, WikiHop, and the Wikidata-answerable subsets of TriviaQA and NaturalQuestions.

In the legal domain, structured knowledge representations have shown promising results for various tasks. Louis et al. (2023) leveraged graph neural networks to enhance statutory article retrieval, demonstrating how graph-based representations can capture complex relationships between legal provisions and improve retrieval performance. Their approach models legal codes as interconnected networks rather than isolated documents, leading to more contextually relevant results. Notably, their work relies on hierarchical legal knowledge structures where statutes are organized in a tree-like manner with clear parent–child relationships.

Building on retrieval approaches, Santosh et al. (2024a) introduced QABISAR, a novel framework utilizing query-article bipartite interactions for statutory article retrieval. This method explicitly models the interactions between query elements and document components, creating more nuanced representations of legal information needs. In a subsequent work, Santosh et al. (2024b) developed CuSINeS, a curriculum-driven structure-induced negative sampling technique that further improves statutory article retrieval by carefully selecting challenging negative examples based on legal document structure. Both approaches leverage the inherent hierarchical organization of legal codes.

For jurisdiction-specific applications, Paul et al (2022) proposed LESICIN, a heterogeneous graph-based approach for automatic legal statute identification from Indian legal documents. Their framework captures the unique structure of the Indian legal system, creating a specialized knowledge graph that significantly outperforms text-based retrieval methods for statute identification. Like other approaches, LESICIN primarily models the legal domain through traditional hierarchical relationships, where higher-level legal principles govern lower-level implementations.

Despite these advances, significant challenges remain in the legal domain. While general knowledge graphs like those explored in Ji et al. (2021) and Sun et al. (2021) have made progress on common-sense reasoning, legal reasoning requires highly specialized domain knowledge structures. Current approaches to legal knowledge graphs predominantly employ hierarchical structures that struggle to capture the dynamic nature of legal interpretations, jurisdictional variations, and the complex interplay between statutes and case law. These hierarchical approaches, while effective for representing formal legal structures, often fail to model the nuanced, non-hierarchical relationships that exist in real-world legal reasoning. Our work addresses these gaps by developing more comprehensive and adaptable knowledge graph structures that go beyond simple hierarchies, specifically designed for complex legal reasoning tasks that involve multidimensional relationships between legal concepts.

## 3 Dataset

IRAC provides a comprehensive problem-solving framework for legal professionals. It takes four stages to transform facts acquired from a legal scenario into legal conclusions: (i) identifying legal issues, (ii) determining the legal rules and precedents pertinent to the issues, (iii) performing analysis by applying the law to the facts and the issues, which requires strong legal reasoning skills, and (iv) drawing conclusions based on the analysis. Legal reasoning is *defeasible* such that there are often more than one reasoning traces leading to the same conclusion or different plausible reasoning traces lead to different reasonable conclusions (Billi et al. 2021). Given a legal scenario, legal professionals' concern is not just about the final conclusion, but also why the conclusion is drawn. Therefore, it is essential to build automatic IRAC analysis tools that produce outcomes for each stage and help them identify any missing reasoning steps, and suggest alternative analysis, when necessary.

While LLMs demonstrate a great potential to automate IRAC analysis in the absence of supervised training data, they suffer from three key limitations: i) wrong references to statutes and precedents, ii) weak legal reasoning capability, and iii) difficulties in filling the gaps between legalese and everyday language. In contrast, prior studies demonstrate the effectiveness of utilizing Retrieval Augmented Generation (RAG) and neuro-symbolic approaches with knowledge bases to enhance the factuality and reasoning capability of LLMs (Gao et al 2023). Therefore, LegalSemi builds the *very first* SKE as a legal knowledge base to facilitate research on neuro-symbolic approaches for legal reasoning and provides an annotated corpus to evaluate system outcomes for each stage of IRAC.

### 3.1 Curation of structured knowledge bases for LegalSemi

Recent advances indicate that it is possible to mitigate the hallucination problem of LLMs and enhance the factual accuracy of their responses by incorporating a knowledge base. While our approach leverages the hierarchical organization of concepts and legal texts, it goes beyond mere hierarchical retrieval. Unlike a simple hierarchical representation of statutes, our SKE captures semantic relationships between legal concepts, rules, and case law. The symbolic components include well-defined relationships between legal concepts and their governing rules, while maintaining the natural language text of legal principles to preserve nuance. This hybrid structure allows for precise rule retrieval while maintaining the interpretative flexibility essential in legal reasoning. For instance, when connecting the concept' invitation to treat' to relevant precedents and statutes, the SKE not only identifies hierarchical relationships but also captures the semantic reasoning patterns across different areas of law, enabling the system to generalize principles across distinct legal domains. These approaches are considered neuro-symbolic because KGs essentially implement the principles of description logic (Baader et al. 2017).

Malaysian Contract Law is taken as the target area of law due to the importance of contracts in everyday life. The corresponding SKE is automatically constructed from the textbook "Law for Business" (Trakic et al. 2022), the Contracts Act 1950 (the primary legislation governing contracts in Malaysia), and 76 court cases pertinent to contracts downloaded from Malaysia e-judgement.[3] It is easy to implement rules to extract legal knowledge from legal documents because the layout of a legal document often resembles the structure of legal knowledge, as evident by the screenshots of the textbook in Appendix 3.

Legal concepts serve as the building blocks of legal doctrine and often act as bridges that connect related knowledge from diverse sources. For example, under the Contract Act 1950, Section 2(a) states: *"when one person signifies to another his willingness to do or to abstain from doing anything, with a view to obtaining the assent*

[3] e-Judgement: https://cms2.kehakiman.gov.my/CommonWeb/ejudgment/SearchPage.aspx?JurisdictionType=ALL

*of that other to the act or abstinences he is said to make a proposal;"*. This section is linked to paragraph P4-014 in the textbook via the legal concept *"offer"*.

The SKE is organised following the structure of the textbook and enriched with the statutes from the primary legislation. The index of the book organises the key concepts of contract law hierarchically, as illustrated in Appendix 3. We extract those concepts from the index and annotate them as nodes at the corresponding levels, such as *main concept* and *subconcept*. The children nodes are linked to their parent nodes using the relation *subconcept of*. Additionally, we represent each chapter title as a node, indicating specific aspects of the Contract Act 1950, such as *Void Agreements*.

Furthermore, the titles, section titles, and etc. were extracted from the Contracts Act 1950. Each title and section is then represented as a node in the SKE. Then, several relations were introduced to associate the nodes derived from the book with the relevant ones in the legislation. For instance, each chapter is associated with the relevant sections of the legislation. Figure 2 shows the snippet structure of the SKE.

To bridge the language gap between legalese and plain English, interpretations were also extracted from the book that provides layman explanations of the corresponding statutes in the legislation. Each interpretation is represented as a node in the SKE, and the *mentions* relation is used to link an interpretation to the relevant statute. Overall, the SKE comprises 3,207 nodes and 2,441 edges, stored in a graph database

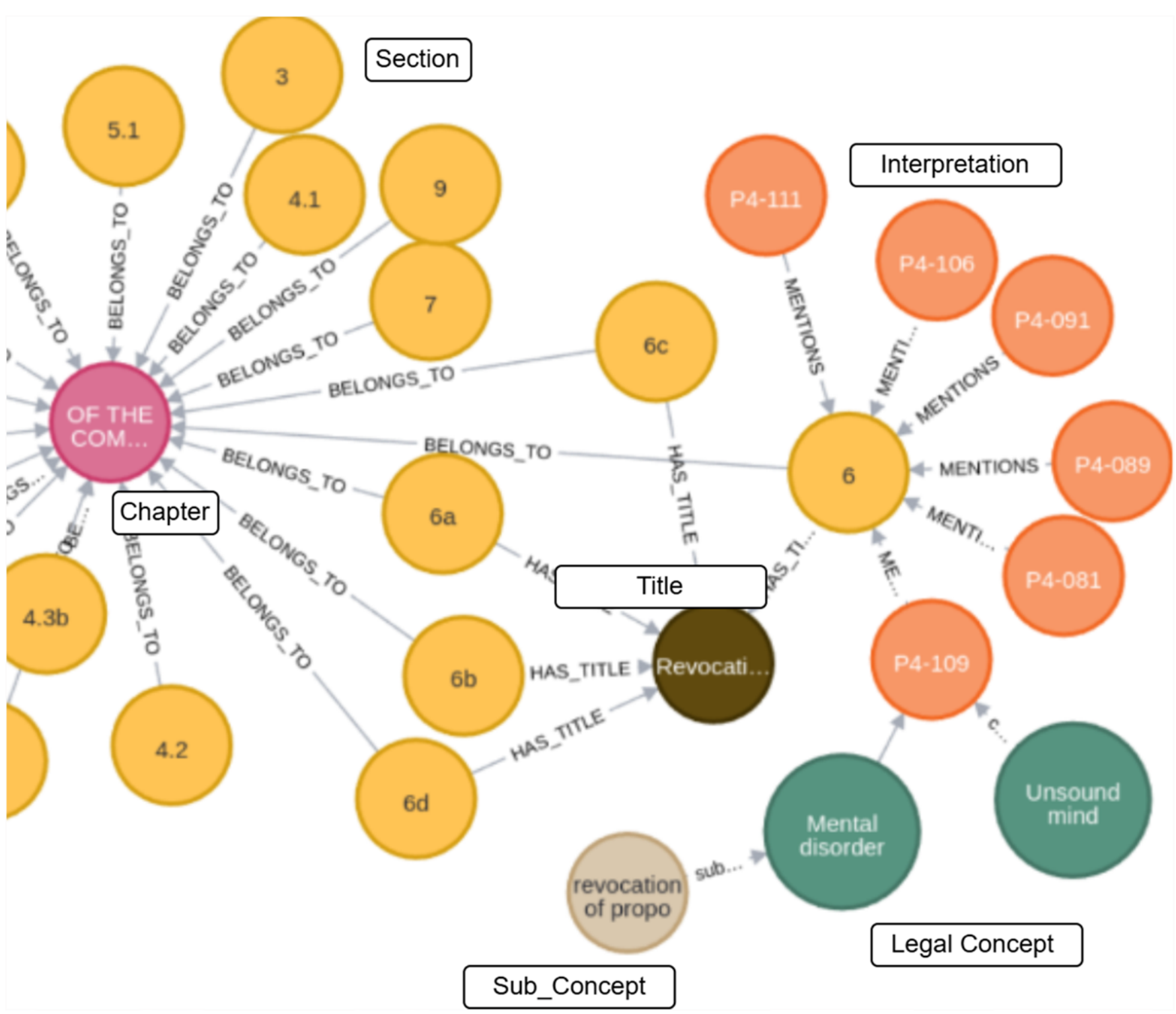

**Fig. 2** A snippet of the SKE. The circles represent nodes, and the lines indicate the relationships between these nodes. Detailed descriptions of the nodes are provided in Table 1

system, namely Neo4j for easy data exchange. Further details about the SKE can be found in Table 1.

### 3.2 Corpus construction

From a pool of applicants from Malaysia, we carefully selected six data annotators to assist with scenario selection and annotating IRAC analysis. This annotator team comprises four second-year law students from two Malaysian universities, as well as two junior lawyers. Out of the six annotators, one of the law students and the two junior lawyers formed the expert team, given their vast knowledge and experience. The annotated corpus comprises 54 legal scenarios covering five chapters with 55 subtopics in a textbook of Malaysian Contract Law. Each scenario reflects real-world legal problems. While the rigorous annotation task takes around three hours per scenario for the IRAC analysis, our easy-to-use annotation tool[4] significantly boosts annotators' productivity.

#### 3.2.1 Scenario selection

To ensure diversity of scenarios and coverage of legal concepts pertinent to *formation of contracts*, we gather scenarios based on the law textbook "Law for Business" (Trakic et al 2022). The book is used by business students when studying business law, and it includes the main aspects of contract law. In particular, we choose five main topics: *offer and acceptance*, *consideration*, *certainty*, *capacity*, and *intention to create legal relations*. The corresponding chapters in the textbook are Chapter 4"Formation of Contract: Proposal and Acceptance", and Chapter 7"Intention to Create Legal Relationships and Capacity". The section headings of these chapters represent the corresponding subtopics, such as *proposal*, *acceptance*, and *minors* etc.. There are 55 unique subtopics extracted from the textbook subheadings.

First, we asked two annotators to create 24 scenarios which were modified from tutorial questions, books, and past examination questions. Next, for the remaining subtopics, we utilised GPT-3.5 turbo to suggest 30 candidate scenarios with the prompt *"You are a legal professional, based on the example scenarios, main topic, and subtopics, create a new scenario around average length"*. To ensure the quality of the scenarios, the two junior lawyers evaluated all 54 scenarios using the questions shown in Fig. 3 and made the necessary revisions. Lastly, the expert law student double-checked all the revisions done by the two junior lawyers. The average length of the scenarios is 800 words.

**Scenario quality analysis** As shown in Table 2, the human evaluation of contract law scenarios created by ChatGPT revealed varying degrees of coverage and quality across multiple dimensions. We adopted a hybrid approach combining human-created and ChatGPT-generated scenarios to balance quality and resource efficiency. This methodology represents an optimal solution within budget and time constraints, as creating all scenarios manually would have been prohibitively expensive and time-

[4] Annotation Platform Example: https://christinakang.github.io/dataAnnotationPlatform/Example.html

**Table 1** The table summarises the SKE implemented in Neo4j, detailing node counts by label and relationship counts by type. This structure organizes contract law data, including sections, interpretations, legal concepts, and their extensions, with relationships linking chapters and titles

| Category | Details | Count | Example/Notes |
|---|---|---|---|
| Nodes | Total: 3207 | | |
| Contract law section | Represents sections of contract law. | 304 | Section 5.2, "An acceptance may be revoked at any time before the communication…" |
| Interpretation | Interpretations of contract law extracted from book content, labeled with paragraph IDs. | 1623 | "If a statement does not satisfy the elements of a proposal, it may be an invitation to treat…" |
| Extend content | Footnotes or extensions related to interpretations. | 307 | "Partridge v Crittenden [1968] 1 WLR Facts: The defendant advertised the…" |
| OtherCase | Represents other cases with semi-legal context. | 59 | Hughes v Metropolitan Railway Co |
| legalConcepts | Semi-legal concepts related to legal principles. | 450 | Communication. Main Concepts:224. 49 sub concept and 177 sub sub concepts. |
| Relationships | Total: 2441 | | |
| CASE_CONCEPT_OF | Connects case and main concept | 627 | Case_name: Hughes v Metropolitan Rly Co → concept: promissory estoppel |
| CONCEPT_OF_INTERPRETATION | Connects the main concept and interpretation | 384 | Acceptance concept_of Under section 3, an acceptance can be communicated by doing something (act) or by refraining oneself from doing something (omission). Either way, it… |
| ADDITIONAL_INFO_OF | Connects the interpretation and extended content | 307 | Hughes v Metropolitan Rly Co (1877) 2 App Cas 439 (House of Lords) Facts: In October 1874, a six-month notice was given by the lessor to the lessee … Related to One of the earlier cases where the doctrine of promissory estoppel was applied and explained, although in a much wider context of estoppel in equity… |
| HAS CHAPTER | Connects a section to its chapter | 304 | Chapter title: "Of Contracts, Voidable Contracts and Void Agreements." |
| HAS TITLE | Connects a section to its title | 304 | Title: "What considerations and objects are lawful, and what not." |

**Table 1** (continued)

| Category | Details | Count | Example/Notes |
|---|---|---|---|
| MENTIONS_SECTION | Connects the interpretation and section | 222 | The definition of "agent" and "principal" is provided in section 135 of the Contracts Act 1950. An "agent" is defined as "a person employed to do any act for another or to represent another in dealings with third persons" whereas the "principal" is defined as "the person for whom such an act is done, or who is so represented". The agency relationship is depicted in Figure 4 as follows: mentions: chapter_title Appoinment and Authority of Agents content An "agent" is a person employed to do any act for another or to represent another in dealings with third persons. The person for whom such act is done… Show all section_id 135 title "Agent" and "principal" |
| SUB_SUB_CONCEPT_OF | Connects the sub sub concept and sub concept | 155 | Third party sub sub concept of communication |
| SUB_CONCEPT_OF | Connects the main concept and sub concept | 138 | Communication sub concept of proposal revocation |

intensive, while relying solely on AI-generated content would risk quality issues for complex legal concepts. For the ChatGPT-created scenarios, we employed in-context learning (Brown et al. 2020; Shu et al. 2024) by providing examples of high-quality human-created scenarios as prompts, which significantly improved generation quality. Most of the main topics and subtopics were covered in the created scenarios when we included partial coverage, though gaps remain compared to human-created scenarios. The statistics demonstrate that 93.1% of scenarios were rated as coherent (logical and comprehensive), indicating strong overall quality despite content gaps in some cases. For scenarios requiring improvement, we implemented a two-stage review process where annotators first evaluated each ChatGPT-generated scenario against established criteria, then revised those that missed the target topic, with a second annotator reviewing these changes to ensure quality control.

### 3.2.2 IRAC annotations

For IRAC annotations, the legal concepts act as a bridge, connecting the facts in a scenario to the professional legal knowledge, including statutes and precedents. We adopted the IRAC analysis based on the annotation guideline in Appendix 1.

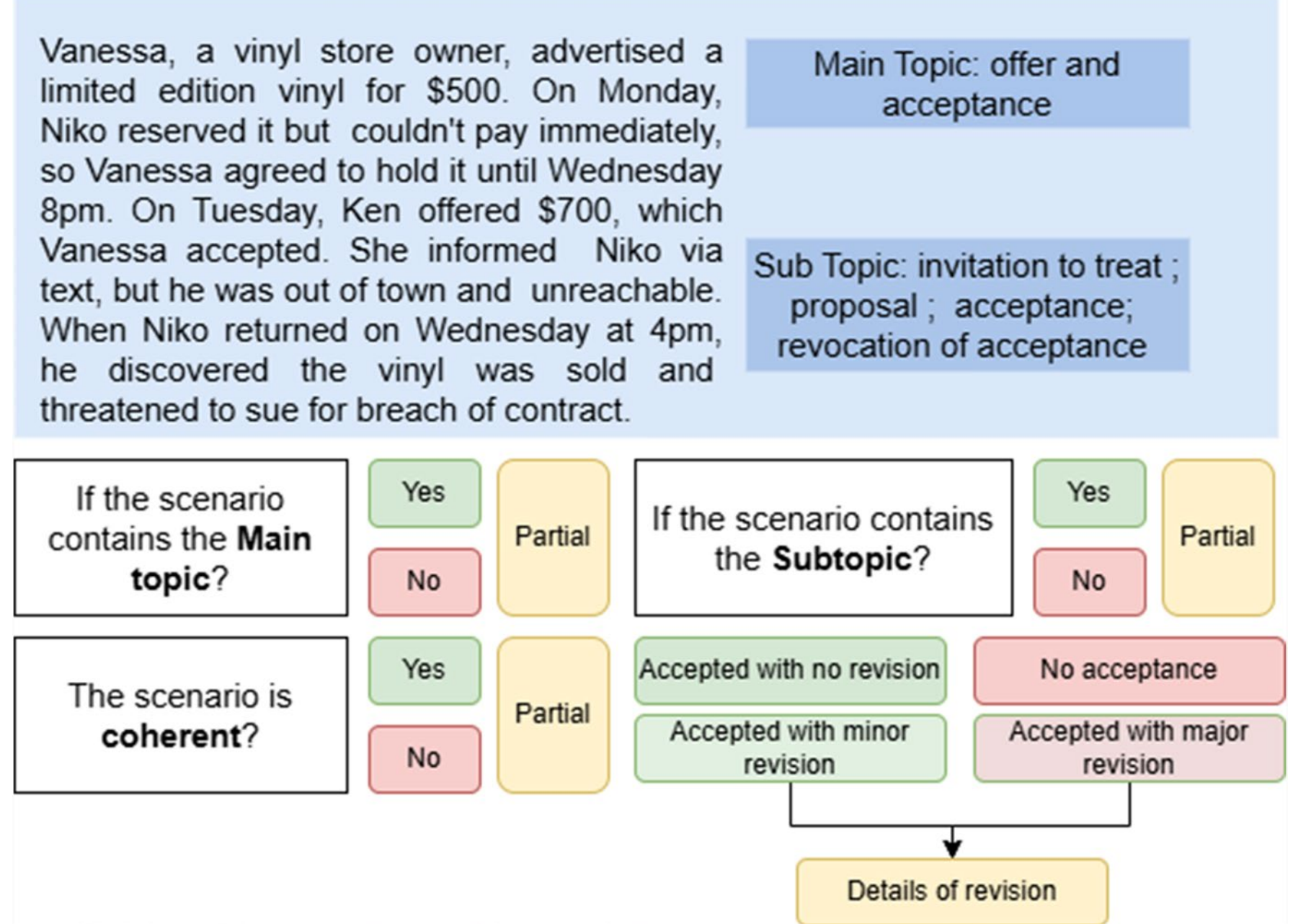

**Fig. 3** A scenario with quality assessment questions

**Table 2** Human evaluation of contract law scenarios

| Evaluation aspect | Category | Percentage (%) |
|---|---|---|
| Main topic coverage | Yes | 41.4 |
| | No | 34.5 |
| | Partial | 24.1 |
| Subtopic coverage | Yes | 55.2 |
| | No | 27.6 |
| | Partial | 17.2 |
| Coherence assessment | Agree | 93.1 |
| | Neutral | 3.4 |
| | Disagree | 3.4 |
| Revision status | No revision | 37.9 |
| | Minor revision | 34.5 |
| | Major revision | 3.4 |
| | No acceptance | 24.1 |

**Legal concepts** Using the legal concepts in the SKE, annotators were asked to identify and highlight relevant legal concepts within a given scenario. If a concept, such as “*offeror*”, is not absent from the textbook, they are allowed to add new concepts into the SKE.

**Issues** A legal issue is a point of dispute involving the interpretation, application, or violation of laws. All six annotators identified issues in scenarios, focusing on whether a valid contract exists between parties. The main challenge in this phase is breaking down the issues into specific questions, such as *"Whether there was an acceptance by Vanessa?"*, as shown in the example scenario (Fig. 1).

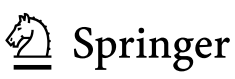

**Rules** A rule specifies the laws applicable to the issues. The annotators are asked to locate the appropriate cases and/or statute sections from the Contract Act 1950 pertinent to issues. For statutory law, the annotation tool offers a drop-down menu to select relevant sections from the 280 sections available in the SKE. For case law, the tool includes text fields for related court cases along with the corresponding page numbers in the cases. For instance, *Eckhardt Marine GMBH v Sheriff, High Court of Malaya, Seremban & Ors [2001] 4 MLJ 4 (CA) [3/4]*. To enable reuse and reference of those rules, the provided cases are displayed as buttons in the user interface so that annotators can refer to those cases by clicking on the buttons (see Fig. 22 in the Appendix 2).

**Application** In the Application section, annotators applied the rules identified in the rules section to the specific facts of the issues, in a given scenario, step by step. They are encouraged to use the conditional statements in the form of "IF…THEN…" to articulate each reasoning step. Figure 4 illustrates the application section of our example. As legal reasoning is defeasible, annotators can make assumptions due to incomplete information. Different assumptions may lead to different conclusions, thus it is essential to discuss and justify these assumptions in the corresponding reasoning step. The application is the most important and challenging section of an IRAC because it develops the answer to the issue at hand.

A discussion with the annotators was carried out to define suitable condition types for LegalSemi. Table 3 lists six condition types defined for the annotation. The most applied condition type in LegalSemi is the IF…THEN… structure.

**Conclusion** The conclusion section directly answers the questions in the issue section, without introducing any new rules and analysis. Following the common practice in law, annotators were asked to write the full sentence of a conclusion, such as *" There is no contract between Vanessa and Niko."*.

#### 3.2.3 Data quality assurance

Given a scenario, there are many plausible IRAC analysis, because different assumptions and different interpretations of rules may lead to different conclusions. As such, IRAC analysis can be very subjective and challenging (Braun 2024). It takes roughly three hours to perform a single IRAC analysis, hence it is infeasible to annotate all possible IRAC analysis. As such, we verified the quality of an IRAC analysis by asking another annotator to act as an evaluator. Specifically, an evaluator can either agree, disagree, or partially agree with an IRAC analysis. The ratio of overall agreements across all 54 scenarios exceeds 0.8, indicating a high level of annotation quality. In instances of disagreement, we consulted an expert to make the final decision and to implement necessary adjustments based on the common methods suggested here (Braun 2024).

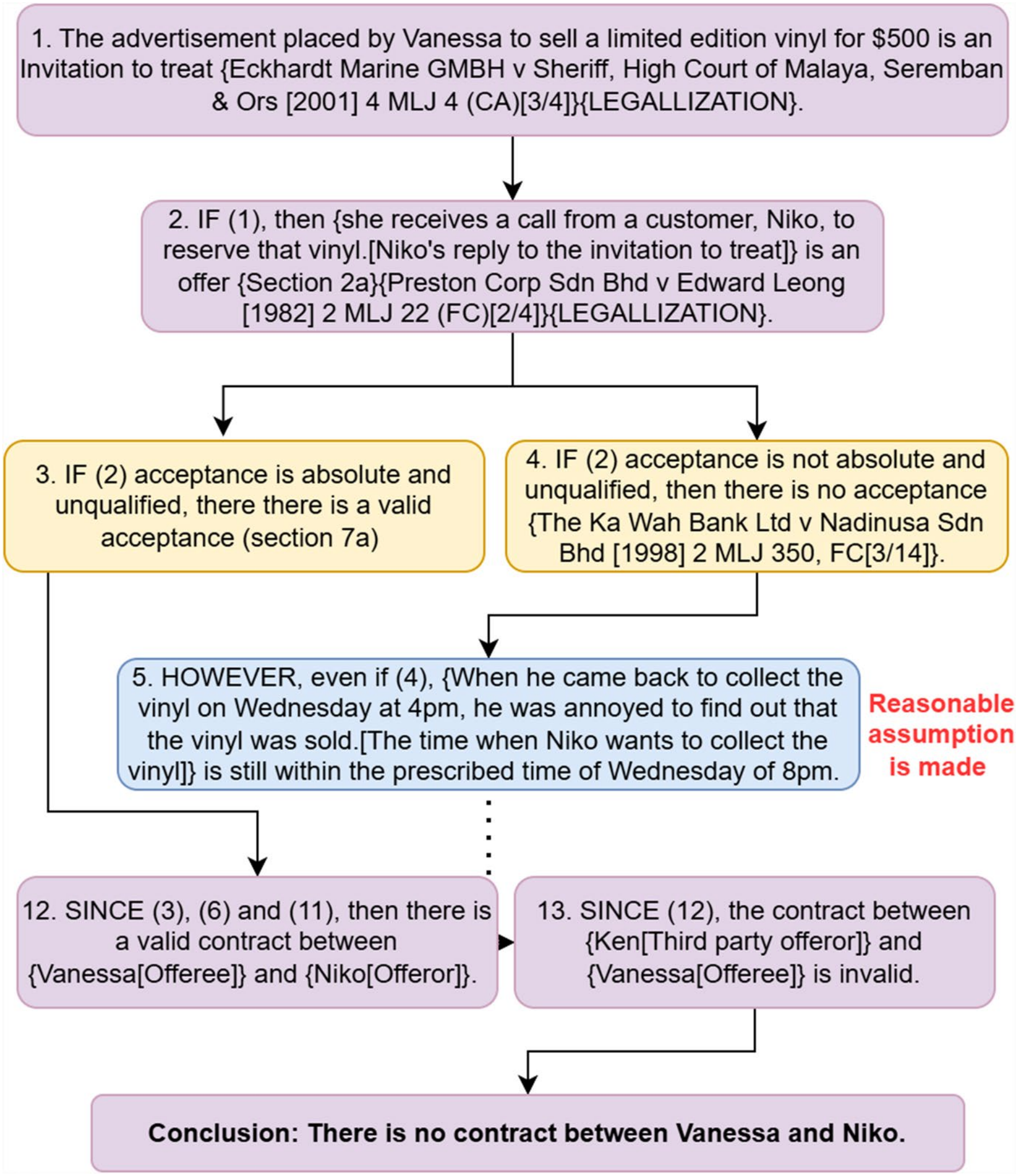

**Fig. 4** An example of the application section

**Annotator quality** IRAC analysis is challenging. Hence, as mentioned above, we selected only annotators who have a strong legal background. The law students were required to have achieved at least a B grade in related law subjects. All annotators must pass a specialized pre-test before being recruited. Financial compensation is MYR30 per hour of annotation work, which is above the minimum wage MYR7.21 in Malaysia, reflecting the complexity and rigour of the annotation tasks.

**Human evaluation results** Human evaluation is conducted for issue identification, application, and conclusion. These aspects require human judgment based on expertise and credentials to ensure accuracy and reliability in the evaluation process.

**Table 3** Application of conditional types

| Condition type | Usage | Example | Count |
|---|---|---|---|
| If… Then | Used to state a condition and its consequence | IF {She placed an advertisement on a social media platform selling a limited edition vinyl for $500 [advertisements: invitation to treat]} is an invitation to treat, THEN the call from a customer, Niko, to reserve that vinyl is an offer | 54 |
| According to… | Used to refer to legal cases, statutes, or authoritative sources | ACCORDING TO {Eckhardt Marine GmbH v Sheriff (2001) 4 MLJ 49(Reimers and Gurevych 2019; Artificial Intelligence (BAAI), B.A. 2023)}, IF {Lowel[offeror]} puts up {advertisement[Advertisements to treat]}, THEN it is not an offer but an invitation to treat | 39 |
| Since… Then | Used to show a reason and its result | SINCE {Lowel responded by saying that 500 is too cheap, and that the lowest price she is willing to sell is RM 700. [Counter offers of initial proposal by proposee]} is not absolute and unqualified THEN there is no valid acceptance | 52 |
| However… If… Then | Used to introduce an exception or a contrasting condition and its consequence | HOWEVER IF the agreement between Penny and Tina is not supported by considerations THEN it will be void even if it is supported by intention to create legal relation | 42 |
| Even if… Then | Used to express a consequence that applies despite a condition | EVEN IF {Nina[Capacity contracting with]} has lied about her age, she would not be estopped from pleading incapacity | 10 |
| Only if… Then | Used to indicate that a consequence will occur solely under a specific condition | IF (7) THEN agreement is supported by intention ONLY IF stated condition is fulfilled. {Confetti Records (A Firm) and others v Warner Music UK Ltd and another [2003] EWHC 1274 (Ch)} | 12 |

We compared the human evaluation results with the auto-evaluation results by examining the rankings of the models' outcomes. Table 4 displays the human evaluation results for all experiments. We calculated Spearman's rank correlation coefficient (Statistics 2023) for each experiment to assess the alignment between human and automated evaluations.

We ranked models based on the 'Pass' rate and compared it to the 'Agree' rate from the automated evaluation matrix. For human evaluation, we used a five-level grading scale: 1 (Fail), 2 (Pass), 3 (Credit), 4 (Distinction), and 5 (High Distinction), and compared these rankings with the 'Agree' rate from the automated evaluation. In conclusion, we compared the entailment rate with the 'Agree' rate from the automated evaluation matrix.

**Table 4** Spearman's rank correlation coefficients

| Criteria | Issues | Application | Conclusion | Human average |
|---|---|---|---|---|
| Coefficient | 0.89 | 0.86 | 0.91 | 0.89 |

**Table 5** LegalSemi with other most relevant datasets

| | No. scenario | Full IRAC | Legal concept | Rules | Annotated application | SKE |
|---|---|---|---|---|---|---|
| SIRAC | 40 | yes | 0 | 58 | Yes | No |
| LegalSemi | **54** | **yes** | **297** | **90** | **Yes** | **Yes** |
| SARA entailment | 277 | no | 38 | 9 | No | No |
| SARA numeric | 100 | no | 38 | 9 | No | No |
| Legal Bench | 59 | no | 0 | 18 | No | No |

From the results in Table 4, the average Spearman's rank correlation coefficient, as indicated by the human average is approximately 0.89. This average further emphasizes the overall strong positive correlation between human and automated evaluations across different criteria.

#### 3.2.4 Summary of the corpus

Our corpus comprises 54 scenarios, 243 issues as decomposed questions, 197 mentions of legal concepts (70 unique), 268 sections of the Contracts Act 1950 (44 unique), 76 court cases, and 607 reasoning paths. On average, each application involves 11.25 reasoning steps to draw a conclusion for the main questions. The most common legal concepts encountered include "offeror", "offeree", and "proposal", reflecting the frequent focus on contract formation. Similarly, the law sections most often cited are s2(a), s2(d), s7(a), s2(e), and s2(b).

#### 3.2.5 Dataset comparison

We compare our corpus with the publicly available corpora in Table 5. Among them, SIRAC (Kang et al. 2023) is the only one that includes annotations of full IRAC analysis for legal scenarios. LegalSemi improves by i) adding a SKE to support neuro-symbolic approaches, ii) introducing annotations of legal concepts to facilitate evaluation of neuro-symbolic approaches, iii) decomposing the main legal questions into scenario-based issues, instead of using fixed issues across scenarios as in SIRAC, and iv) include a test set with longer scenarios, closer to the real world scenarios, that require more complex reasoning.

SARA (Holzenberger and Van Durme 2021) focuses on legal question answering in Taxation Law. It annotates structured reasoning paths involving merely nine rules in total for the QAs but does not include any annotations of IRAC. Legal Bench (Guha et al. 2022) covers diverse legal AI tasks but does not include full IRAC analysis, particularly the detailed analysis in Application.

## 4 Experiments

In this section, we empirically demonstrate the usefulness of LegalSemi for IRAC analysis and highlight the open challenges for future research. To investigate models' performance at each stage and prevent cascading errors, we evaluated each IRAC stage as an NLP task sequentially, using expert-curated outputs from previous stages as inputs for subsequent analysis. For issue generation and rule retrieval, we evaluated to what extent legal concepts can improve the performance of the corresponding tasks, hence we also evaluated the models' ability on identifying legal concepts.

### 4.1 Models details

We evaluated four LLMs in this study: GPT-3.5 turbo, Llama 2, Mistral, and Gemini. This study employs LLMs because they demonstrated superiority over traditional NLP and machine learning approaches across diverse tasks, particularly in zero-shot scenarios where IRAC analysis training data is unavailable. Our comparative evaluation examines each model's performance across three key dimensions: accuracy of legal reasoning, coherence of analysis, and relevance of content.

For GPT-3.5 turbo, we configured the model with a temperature setting of 0.7, a standard choice for GPT-based models that balances creativity with coherence in generated responses. Additionally, the maximum token count was set to 1000, ensuring that the model can provide comprehensive and detailed outputs. Llama 2 70B was chosen for its large parameter size, which was expected to enhance its capacity to grasp nuanced legal details. Mistral 7B, noted for its efficiency and rapid inference capabilities, serves as a contrast to the more parameter-intensive models. Finally, Gemini was included given its prior success in legal text analyses, providing a valuable benchmark for performance. The purpose of this comparison was to identify the most effective model for tasks that require precise legal reasoning and interpretation.

### 4.2 Legal concept identification

Legal concepts play a key role in neuro-symbolic approaches for legal reasoning by linking facts in scenarios with legal knowledge. We investigated how accurate the state-of-the-art (SOTA) LLMs can identify legal concepts in scenarios, because LLMs demonstrate remarkable performance on zero-shot and few-shot learning (Brown et al. 2020). Our prompt for the four LLMs is illustrated in Fig. 5. It begins with instructing the LLM to select relevant concepts from a comprehensive list of concept candidates, followed by providing a scenario and main legal questions. At the end of the prompt, it requires the output format to be a Python list for easy post-processing. To investigate the effectiveness of different structural components of this prompt, we evaluated two simplified variants: i) one omitting the given legal concepts and self-check instruction; and ii) one omitting the self-check instruction.

We utilized a dataset containing 197 legal concepts across all scenarios, with 70 unique concepts identified in our ground truth annotations. We have 224 concepts that were classified as top-level concepts in our legal knowledge database. Our evaluation metrics focus on measuring the LLMs' ability to accurately identify these top-level

**Legal concept identification prompt**

**Instruction**

You are a legal professional. Please identify the legal concepts based on the following scenario in Malaysia, given an example of legal concept identification.

**Given Legal concepts:**

Please select concepts from the following list: [A PYTHON LIST OF CANDIDATE CONCEPTS]

**Input:**

Scenario: [CONTENT OF THE SCENARIO]
Main legal question: [MAIN QUESTION PERTAINING TO THE SCENARIO]

**Self-check instruction:**

Be sure it is reasonable to answer the main legal question.

Instruction

I don't want the full analysis. I just want a list of legal concepts. Please generate the selected concepts in a python list.

**Fig. 5** The prompt template for legal concept identification with the most comprehensive inputs

concepts across the test scenarios. This approach allows us to quantitatively assess the models' capacity for legal concept extraction, which serves as a fundamental component in neuro-symbolic legal reasoning systems.

The performance differences observed across legal concepts may be attributed to multiple factors. Rather than making definitive claims about pre-training data frequency, we acknowledge that the challenges in identifying subordinate legal concepts could stem from their inherent complexity, the context-specific nature of their application, or insufficient definitional context in the prompts provided to the models. Future research could explore enhancing LLM understanding through explicit definitional approaches and contextual enrichment that clarify the relationships between concepts within the Malaysian contract law framework.

**Legal concept evaluation** We evaluated the generated concept lists against ground truth annotations using precision, recall, and F1 score. Precision measures the proportion of correctly identified legal concepts out of all identified concepts. Recall measures the proportion of correctly identified legal concepts out of all relevant concepts in the ground truth, indicating the model's completeness. The F1 score provides a mean of precision and recall, offering a single metric that balances both aspects of accuracy.

Given the hierarchical organization of legal concepts in the index of the textbook "Law for Business" (Trakic et al. 2022), we assessed model performance at two distinct levels of abstraction: top-level concepts (e.g., 'invitation to treat') and their subordinate concepts (e.g., 'audition' and 'advertisement' as specific instances of 'invitation to treat'). This hierarchical evaluation allows for a more nuanced understanding of model performance across different levels of conceptual granularity.

**Table 6** F1 Score for predicting both top-level and subordinate concepts

| Concepts | GPT-3.5 turbo | Llama 2 | Mistral | Gemini |
|---|---|---|---|---|
| Top-Level Concepts | 35% | 34% | 32% | 34% |
| Subordinate Concepts | 8% | 12% | 10% | 11% |

**Issue generation prompt**

**Instruction**

You are a legal professional. Given a legal scenario in Malaysia, generate the decomposed questions step by step based on the hint of the legal concepts.

**Input:**

Scenario: [CONTENT OF THE SCENARIO]
Main legal question: [MAIN QUESTION PERTAINING TO THE SCENARIO]
Legal concepts: [LIST OF LEGAL CONCEPTS FROM GROUND TRUTH]

**Self-check instruction:**

Be sure it is reasonable to answer the main question.

Instruction

In your response, please use a python list for all decomposed questions and sub questions.

**Fig. 6** The prompt template for issue generation

#### 4.2.1 Results

As illustrated in Table 6, all four LLMs performed significantly better at predicting top-level concepts compared to the subordinate ones. For the top-level concepts, GPT-3.5 turbo achieves the highest precision (35%), while Gemini obtains the highest recall (93%). We conjecture that compared to top-level concepts, e.g.”invitation to treat”, the subordinate concepts associate with specific details of contracts, such as”audition” and”advertisement”. Hence, they appear less frequently in the pre-training data of LLMs. This sheds light on the importance of constructing dedicated supervised training data for future research.

### 4.3 Issue identification

Rather than using predefined issue sets as in prior work (Kang et al. 2023; Guha et al. 2022), we evaluated LLMs’ ability to generate issues for each scenario. To investigate the effectiveness of neuro-symbolic approaches, we assessed how identified legal concepts influence the quality of issue generation.

We adopted the same four LLMs as legal concept identification. Their prompt is detailed in Fig. 6. In the prompt, we instructed LLMs to break a main legal question into a list of issues by leveraging relevant ground-truth legal concepts, and asked LLM to self-evaluate its outputs by ensuring they are *reasonable*, as inspired by Hao et al. (2023).

#### 4.3.1 Evaluation details

As issue generation is a language generation task, following (Kang et al. 2023), we apply GPT-3.5 turbo to compare predicted issues with annotated reference issues with a list of criteria detailed in the prompt in Fig. 7. An LLM is expected to select one option from: strongly agree, neutral, or disagree, which is further mapped to a score of 1, 0, and −1, respectively.

To investigate the quality of this automatic metric, we compared the results of the automatic evaluation with those of human evaluation. In the human evaluation, we assessed the quality of an IRAC analysis using a rubric that is widely used in Malaysian contract Law courses (Gerhardt 2008; Carter 2006). The issues of an IRAC analysis receive a *Pass* when they satisfy the corresponding criteria detailed in Appendix D.1, otherwise, they are marked as *Fail*.

We asked two annotators to independently assess the issues generated by two best performing models (Gemini and GPT-3.5 turbo) using three different prompts on 10 randomly selected scenarios. Those prompts differ in whether they omit ground-truth legal concepts and self-check instruction. If there was a disagreement between their assessments, we asked the most experienced annotator with a strong legal background to resolve it. Each generated output marked as *Pass* receives a score of 2, otherwise, a score of 1. We then ranked all LLM configurations according to their average scores and compute the Spearmann rank correlation with the counterpart using the automatic metric. A strong correlation of 0.89 suggests the effectiveness of the automatic metric as shown in Table 4.

#### 4.3.2 Results

Figure 8 depicts the average scores of the automatic metric across all 54 scenarios. Legal concepts significantly enhance performance across all LLMs, with GPT-3.5 Turbo showing the most substantial improvement (28.0% increase). The introduction of legal concepts notably improves issue identification quality across all models. Furthermore, incorporating self-evaluation instructions further enhances performance,

**Evaluation Prompt for Issue Generation**

**Evaluation Criteria:**

You are a legal professional. Based on the evaluation criteria, evaluate the provided decomposed questions compared to the ground truth. Assign scores to each decomposed question indicating its relevance and alignment with the corresponding ground truth question.

Do you agree that the issues (issue and decomposed questions) are Clarity and completeness of the issues and sub-issues? Relevance of the identified issues to the problem at hand.

Appropriateness of the decomposition of complex questions.

Scoring Guidelines:

**1: Strongly agree** - The issue and sub-issues listed in the document are clear, comprehensive, and highly relevant to the problem.

**0: Neutral** - The issues are identified but may require further clarification or additional depth.

**-1: Disagree** - The issues are unclear, incomplete, or not directly related to the problem, according to the document.

**Fig. 7** The evaluation prompt template for issue generation

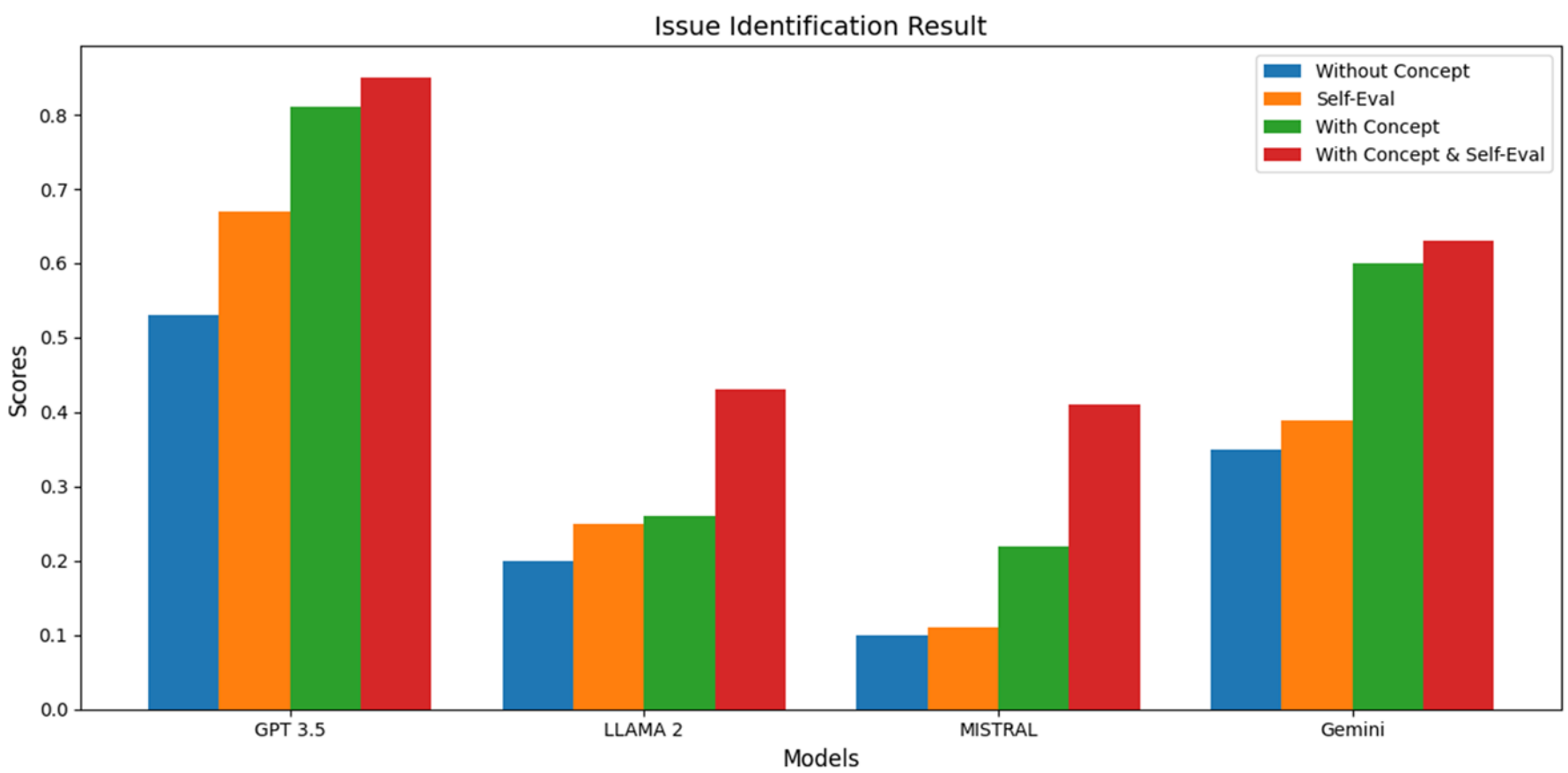

**Fig. 8** The results of issue identification

with the combination of both legal concepts and self-evaluation consistently yielding the best results for all LLMs tested.

Based on our manual inspection, the incorrect issues generated by LLMs are often either irrelevant to the main issue and the facts of a given scenario, or logically not plausible based on relevant legal knowledge. For example, the generated issue "*Did Vanessa breach the contract with Niko by selling the vinyl to Kenn?*" in Fig. 9 is not pertinent to the main issue regarding whether there is a valid contract between the two parties. Another common error is the generation of issues like"*What is the legal concept of "Notice of revocation",* which cannot be considered as an issue.

### 4.4 Rule retrieval

We evaluated the effectiveness of retrieving statutes and cases pertinent to the Contract Act 1950 from our *SKE* using scenarios annotated with legal concepts. This task involves formulating queries based on the scenarios and efficiently ranking relevant statutes and cases above irrelevant ones. A primary challenge lies in bridging the linguistic disparity between the lay language used in scenarios and the formal legal terminology (legalese) present in statutory rules and legal cases. Unlike prior work in legal retrieval that relies on formal legal questions or raw scenarios as queries, we also examined how our SKE mitigates the linguistic gap between legalese and lay language, in particular through illustrations and examples in lay language associated with statutes.

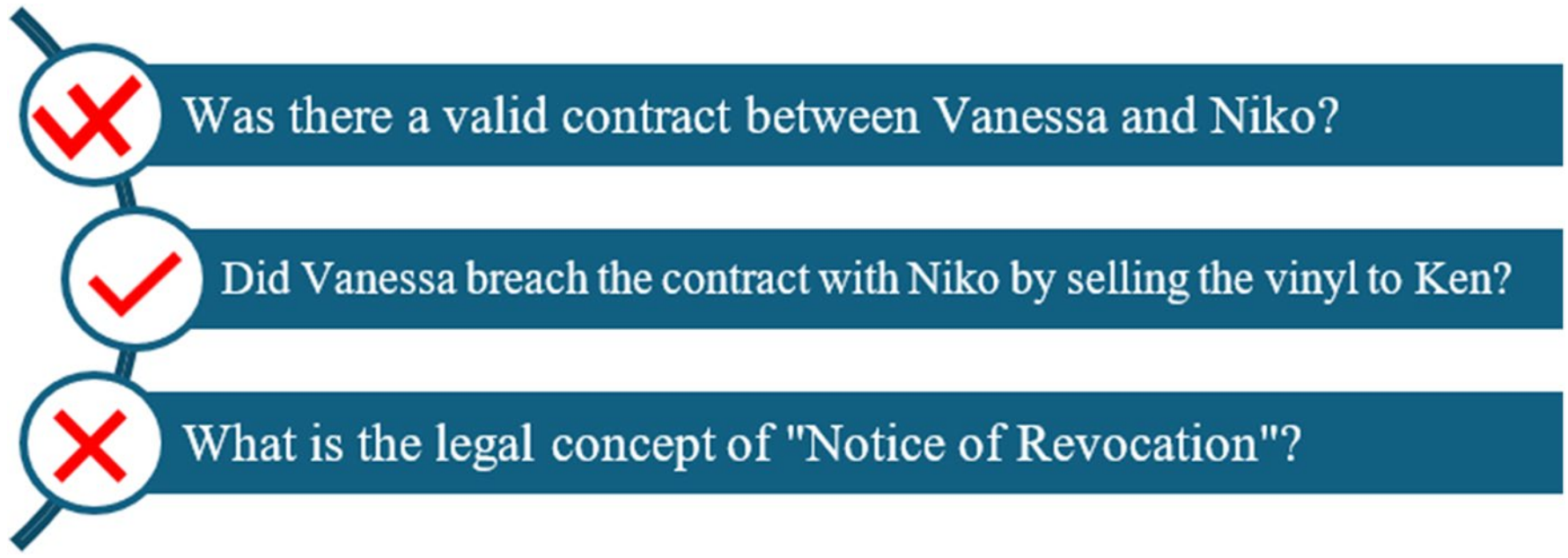

**Fig. 9** The llm output example from the given prompt

To investigate the open challenges associated with this task, we evaluated traditional information retrieval methods, SOTA statute retrieval techniques, and embedding-based approaches. For traditional methods, we employed TF-IDF, BM25, and smooth TD-IDF (Pedregosa et al. 2021),[5] selected for their well-established effectiveness in classical information retrieval tasks. These models are widely recognized for their simplicity, efficiency, and robustness, providing a reliable means of ranking documents based on term frequency and inverse document frequency.

Inspired by the statute retrieval method CASRank (Parashar et al. 2024), which integrates a pre-processing step tailored to statute corpora in India with a ranking algorithm almost identical to Okapi BM25, we enhance the document indexing process by filtering out terms with IDF scores in the lowest 60% and applied Okapi BM25 to compute document relevance scores. This approach, designed to refine the relevance of retrieved documents, is referred to as the *improved BM25*.

Furthermore, we assess embedding-based retrieval models, including BERT (Devlin et al. 2018), Legal-BERT (Chalkidis et al. 2020), Sentence-BERT (Reimers and Gurevych 2019), and the BAAI embedding model (Artificial Intelligence (BAAI), B.A. 2023), chosen for their ability to capture semantic relationships within textual data. These models represent SOTA advancements in retrieval tasks, particularly their capability to address nuanced linguistic challenges such as synonymy and contextual variations. Legal-BERT, specifically fine-tuned on legal texts, is tailored to the intricacies of the legal domain, while Sentence-BERT is optimized for sentence-level semantic similarity. The BAAI embedding model is implemented to capture dense semantic representations across diverse datasets.

### 4.4.1 Methodological details

All evaluated retrieval systems involve constructing queries and indexing statutes and legal cases. They differ in the representations of queries and legal documents, as well as in the similarity functions used to match queries with documents.

**Query** Given a scenario, we consider three variants of an input query: i) the original scenario, ii) a rewritten version of the scenario transformed into a structured list of facts using GPT-3.5 turbo, and iii) legal concepts derived from the scenario. The rewritten facts transform the scenario into precise and actionable statements, while the predicted legal concepts, generated by GPT-3.5 turbo, associate the scenario with the relevant legal knowledge.

[5] This is implemented by enabling smooth idf and sublinear tf parameters in the TfidfVectorizer module of Scikit-Learn.

**Indexing** We compare indexes constructed using varying representations of both statutes and legal cases in the SKE. For statutes, we consider the following representations:

- Original Text: The plain text of individual sections and subsections from the Contract Act 1950. For instance, "when one person signifies to another his willingness to do or to abstain from doing anything, with a view; to obtaining the assent of that other to the act or abstinence, he is said to make a proposal", as stated in Section 2(a).
- Enriched Representation: A combination of the original text with additional metadata, including illustrations, examples, and titles. The illustrations provide explanations of statutes in lay language, while the examples enhance accessibility by offering practical applications of the statutes. As the textbook contains only a limited number of illustrations and examples, we generate them by utilizing GPT-3.5 turbo. The prompt guiding this process is outlined in Fig. 10, and an example of the enriched representation is shown in Table 7.
- Legal Concepts: Top-level legal concepts associated with each statute to capture the semantic essence of the legal provisions.

For legal cases, in addition to indexing the legal concepts associated with each case, we construct an index based on the facts of the cases, as these details closely align

Instruction and Example

**Instruction**

Please provide an explanation of a given law with an illustration and a real-life example, structured as a Python list containing elements for the law, illustration, and example. The output should include:

**Illustration:**

A simplified explanation of the given law in common English and further clarification.

**Example:**

A real-life scenario demonstrating the application of the law.

**Given Law:**

"A 'bailment' is the delivery of goods by one person to another for some purpose, upon a contract that they shall, when the purpose is accomplished, be returned or otherwise disposed of according to the directions of the person delivering them. The person delivering the goods is called the 'bailor'. The person to whom they are delivered is called the 'bailee'."

**Example Output:**

Illustration: "Every person has a right to enter into any type of contract they decide to be appropriate for them. However, there are some circumstances where the law limits the ability of a party to enter into contracts. The reason for such limitation usually is based on policy. The requirement of capacity may be found in section 10(1) of the Contracts Act 1950, which provides:",

Example: "A 16-year-old high school student attempts to take out a loan for a large sum of money to start a business. The bank, upon discovering the student's age, declines to process the loan application, citing the student's lack of legal capacity to enter into such a contract according to the Contracts Act 1950. This act protects minors from being bound by contracts that they may not fully understand or that may not be in their best interest."

**Fig. 10** The prompt for generating illustrations and examples using in-context learning

**Table 7** Examples of interpretation and real-life application

| Section ID | Content | Interpretation | Real-life example |
|---|---|---|---|
| s 2a | When one person signifies to another his willingness to do or to abstain from doing anything, with a view to obtaining the assent of that other to the act or abstinence, he is said to make a proposal | A bailment is the transfer of possession and control of personal property (goods) from one person (the bailor) to another person (the bailee) for a specific purpose, with the understanding that the goods will be returned to the bailor or otherwise disposed of according to their instructions once the purpose is fulfilled | John wants to sell his car and approaches his friend Mark with an offer. John tells Mark that he is willing to sell his car for $10,000. In this scenario, John is making a proposal by expressing his willingness to sell his car at a certain price. Mark can either accept or reject the proposal |
| s 2c | The person making the proposal is called the”promisor” and the person accepting the proposal is called the ”promisee.” | In a legally enforceable promise, there are two parties involved. The person who makes the offer is known as the ‘promisor’ and the person who accepts the offer is referred to as the ‘promisee.’ | In a real-estate transaction, the seller proposes to sell their property to the buyer at a specific price. The buyer accepts this offer, resulting in a legally binding contract between them. In this scenario, the seller is the ‘promisor’ because they made the offer, and the buyer is the ‘promisee’ because they accepted the offer |

with the facts in a scenario. Each type of representations is indexed using both the aforementioned traditional retrieval methods (TF-IDF and its variants) and embedding-based models. For embedding models, each type of vector representations is generated using the corresponding pre-trained model, such as bert-base-uncased for BERT and BAAI/bge-large-en-v1.5 for BAAI, with cosine similarity applied for retrieval through FAISS (Jegou et al. 2022).

Algorithm 1 shows how we evaluated each type of representation for statute and case retrieval. When we applied predicted concepts during this process, we selected top-*k* concepts from the index, followed by identifying the statutes or cases linked to those concepts.

**Algorithm 1** Legal Rule Retrieval Framework

**Input:** Legal scenario content $C$, rewritten facts $F$, predicted legal concepts $L$
**Index Types:** *section original*, *section combined*, *case facts*, *concepts index*
**Output:** Retrieved results $R$ for sections and cases for each query, index, and top-$k$ value
1: **for** each query $q \in \{C,F,L\}$ **do** > Scenario, Facts, Legal Concepts
2: **for** each index type $I \in \{$*section original ,section combined, case facts*$\}$ **do** > Statute and Case Indexes
3: **for** $k \in \{5,10,20\}$ **do** > Top-$k$ Values
4: Compute similarity between $q$ and $I$ using retrieval models (e.g., BM25, embeddings)
5: Retrieve top-$k$ sections or cases
6: Store results $R_{q,I,k}$ for evaluation
7: **end for**
8: **end for**
9: **end for**
10: **for** each query $q \in \{F,L\}$ **do** > Facts and Predicted Legal Concepts
11: **for** $k \in \{5,10,20\}$ **do** >Top-$k$ Values
12: Compute similarity between $q$ and *concepts index*
13: Retrieve top-$k$ legal concepts
14: Use Neo4j to find:
Related statute sections with *section hop* $\in \{1,4\}$
Related court cases with *case hop* $\in \{1,2\}$
15: Store results $R_{q,concepts,k}$ for evaluation
16: **end for**
17: **end for**
18: **Return** $R$: Results for all query-index combinations and top-$k$ values

### 4.4.2 Evaluation metrics

We evaluated the performance of our retrieval framework using four widely adopted metrics in information retrieval: Precision, Recall, F1-score, and Mean Average Precision (MAP). These metrics collectively assess the quality of retrieval from different perspectives, as detailed below:

- **Precision@k:** It measures the fraction of retrieved items that are relevant to the query. For a given top-$k$ result set, it is computed as:

$$Precision@k = \frac{Number\ of\ Relevant\ Items\ Retrieved\ in\ the\ Top-k\ List}{k}$$

Higher precision indicates that the retrieval system successfully prioritizes relevant documents among the top-ranked results.

- **Recall@k:** It evaluates the fraction of relevant items that are successfully retrieved in the top-$k$ list:

$$Recall@k = \frac{Number\ of\ Relevant\ Items\ Retrieved\ in\ the\ Top-k\ List}{Total\ Number\ of\ Relevant\ Items}$$

High recall ensures the system retrieves a significant proportion of the relevant documents, even at the expense of including some irrelevant ones.

- **F1@k:** It combines Precision@k and Recall@k into a single harmonic mean to balance both metrics:

$$F1@k = 2 \cdot \frac{Precision@k \cdot Recall@k}{Precision@k + Recall@k}$$

This is particularly useful when both precision and recall are important for the retrieval task.

- **Mean Average Precision (MAP):** MAP assesses the quality of ranking by computing the average precision at each rank position where a relevant item occurs and averaging this value across all queries:

$$MAP = \frac{1}{|Q|}\sum_{q \in Q} \frac{\sum_{k=1}^{n} P(k) \cdot relk)}{Total\ Relevant\ items\ for\ q}$$

Here, $P(k)$ is Precision at rank $k$, and rel($k$) is a binary indicator of relevance at rank $k$.

These metrics, inspired by established evaluation frameworks (Parashar et al. 2024; Shao et al 2021), enable a comprehensive analysis of retrieval performance across multiple dimensions, including ranking quality, relevance, and completeness.

#### 4.4.3 Results and discussion

**Statute retrieval** Figure 11 illustrates the statute retrieval results in terms of Precision@10, Recall@10, and F1@10, across all evaluated IR models, using facts as queries and enriched representations in the indexes. The TF-IDF variants consistently outperform the embedding based approaches, indicating that embeddings may inadequately increase linguistic gap between queries and relevant statutes. Smooth TF-IDF achieves superior performance over all the other IR models across different types of queries and indices. It achieves the best precision (14.94%) at top-10 and the highest recall (60.72%) at top-20 with facts as queries and the index constructed with enriched representation.

**Case retrieval** As illustrated in Fig. 12, the best performing model for case retrieval is still smooth TF-IDF, which achieves a precision of 20.90% at top-5 and a recall of 30.94% at top-20, using scenarios as queries and an index constructed with enriched representations. The same as statute retrieval, indices with enriched representation perform significantly better than their alternatives, while scenarios provide more comprehensive information as queries compared to the counterparts.

**Choice of queries and indices** Figure 13 illustrates the F1@10 scores achieved using smooth TFIDF across various types of queries and indices. The results clearly show that indices with enriched representations perform best for both statute and case retrieval. The inclusion of illustrations and examples significantly enhances retrieval quality by providing additional contextual information, enabling queries to align more effectively with the indexed content and yielding more accurate and relevant results. Legal concepts did not perform well largely because the coverage of annotated legal concepts for statutes and cases is too low. Among query types, facts are most effective for statute retrieval, while scenarios perform best for case retrieval. These findings are consistent across other IR models as well.

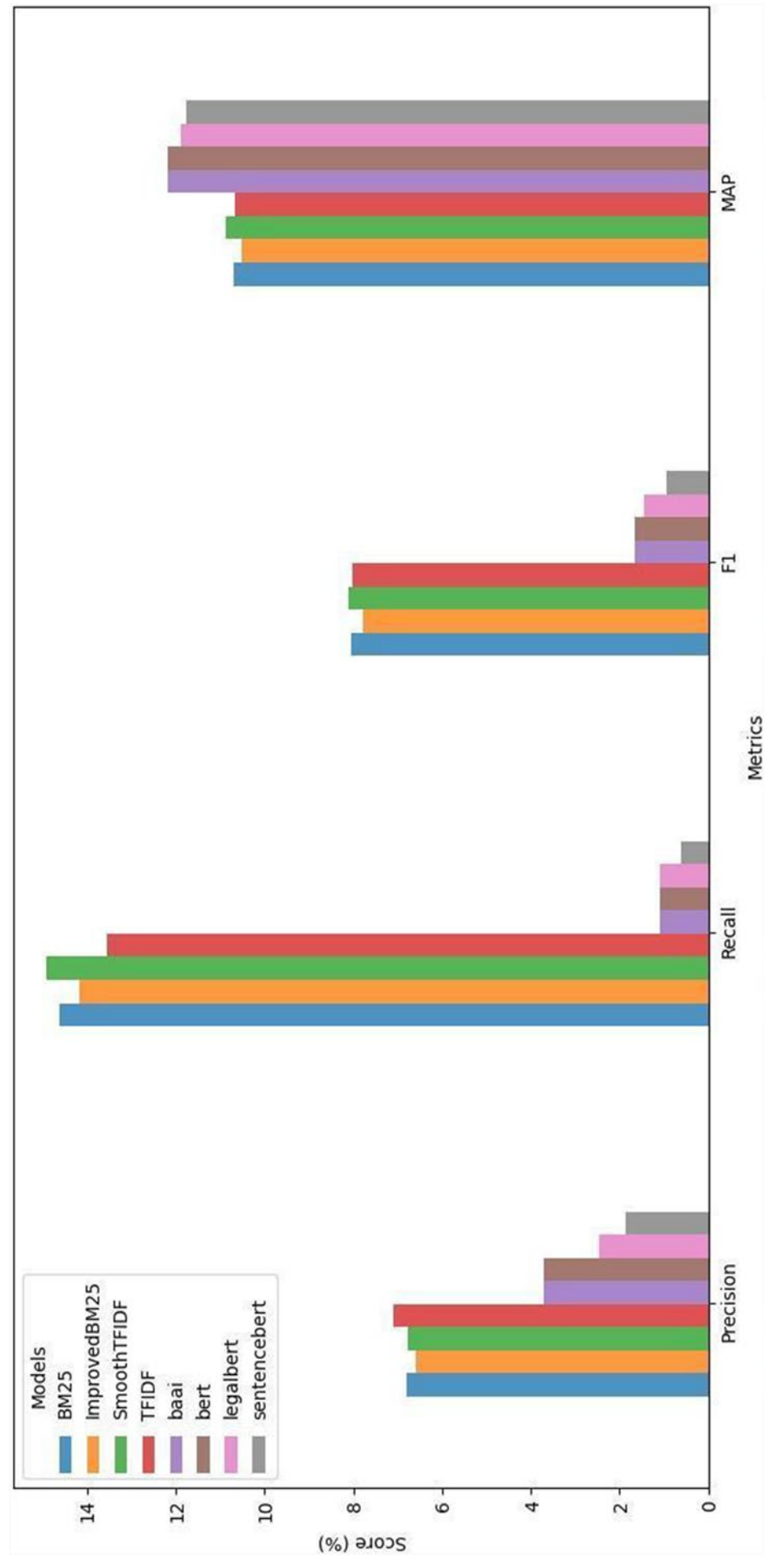

**Fig. 11** Statute retrieval results in terms of Precision@10, Recall@10, F1@10 and MAP

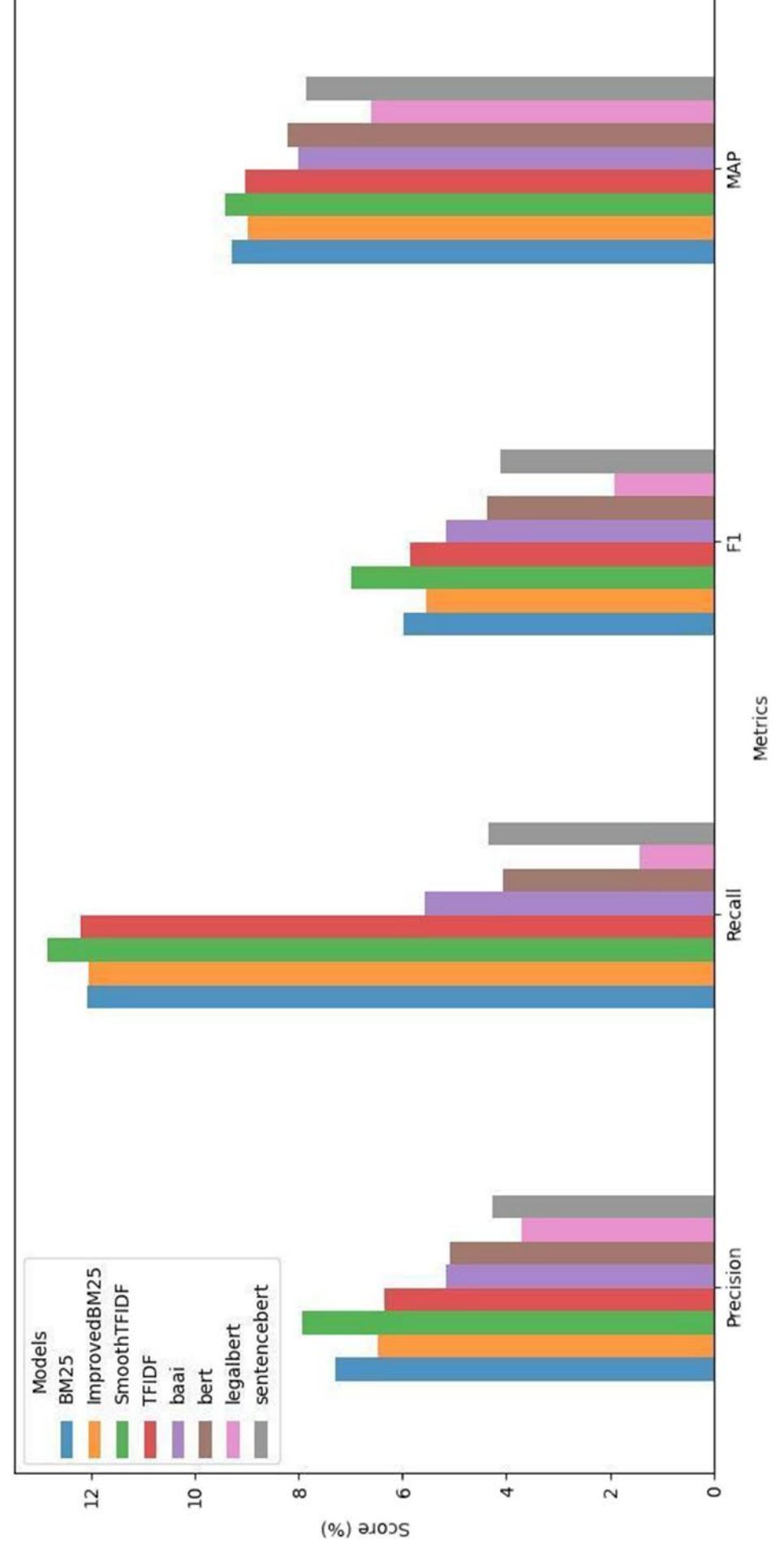

**Fig. 12** Case retrieval results in terms of Precision@10, Recall@10, F1@10 and MAP

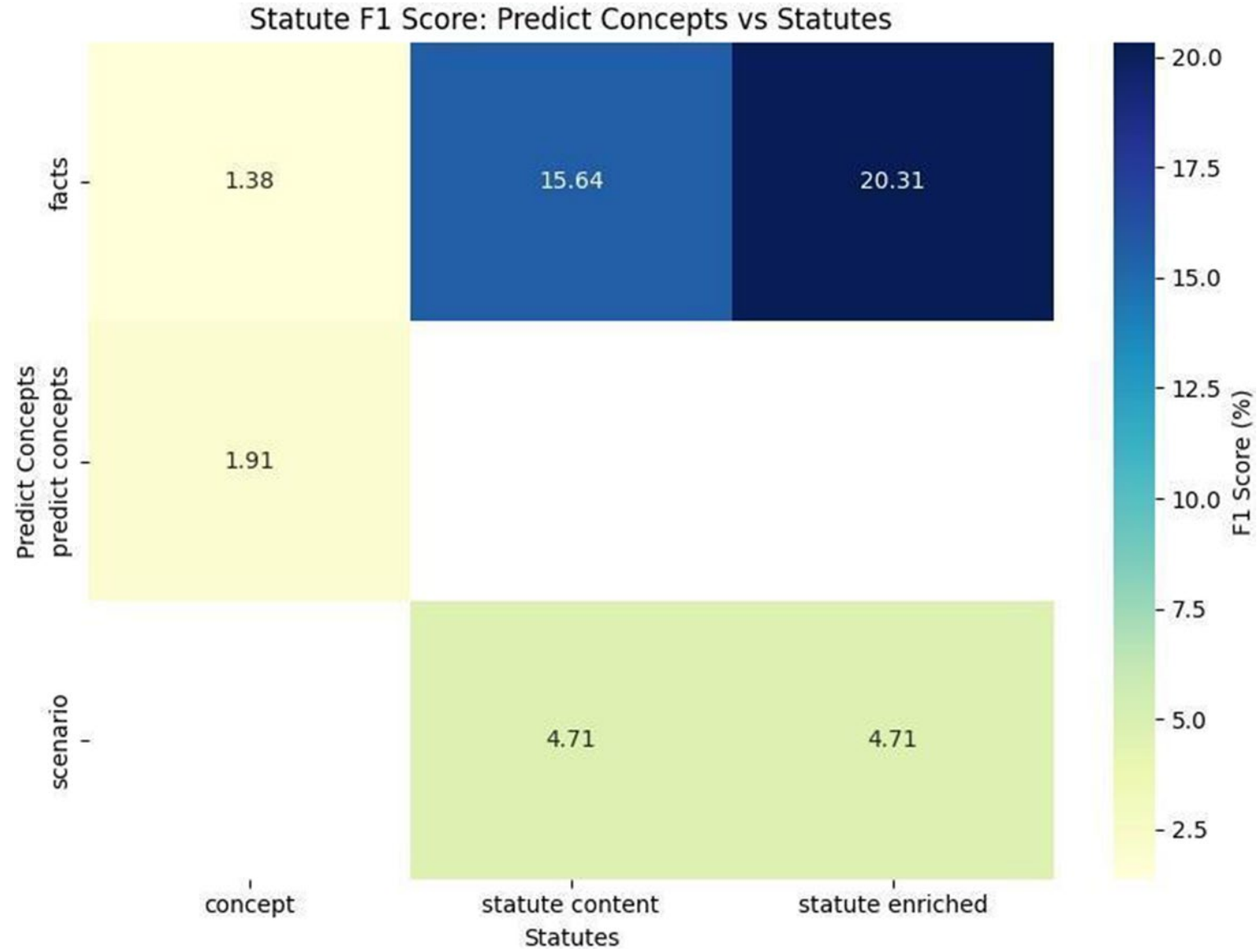

**Fig. 13** Impact of Input Queries on F1 Scores for Statute (Smooth TF-IDF, Top-K = 10)

**TopK analysis** Figure 14 illustrates the Top-K results achieved by Smooth TF-IDF for both statute and case retrieval. While top-20 results yield higher recall, particularly in case retrieval, it often comes at the cost of reduced precision. Precision remains relatively stable across different TopK values, as our evaluation algorithm leverages legal concepts to associate initial retrieval results with the statutes or cases linked to those concepts, leading to a final ranking list larger than *k*.

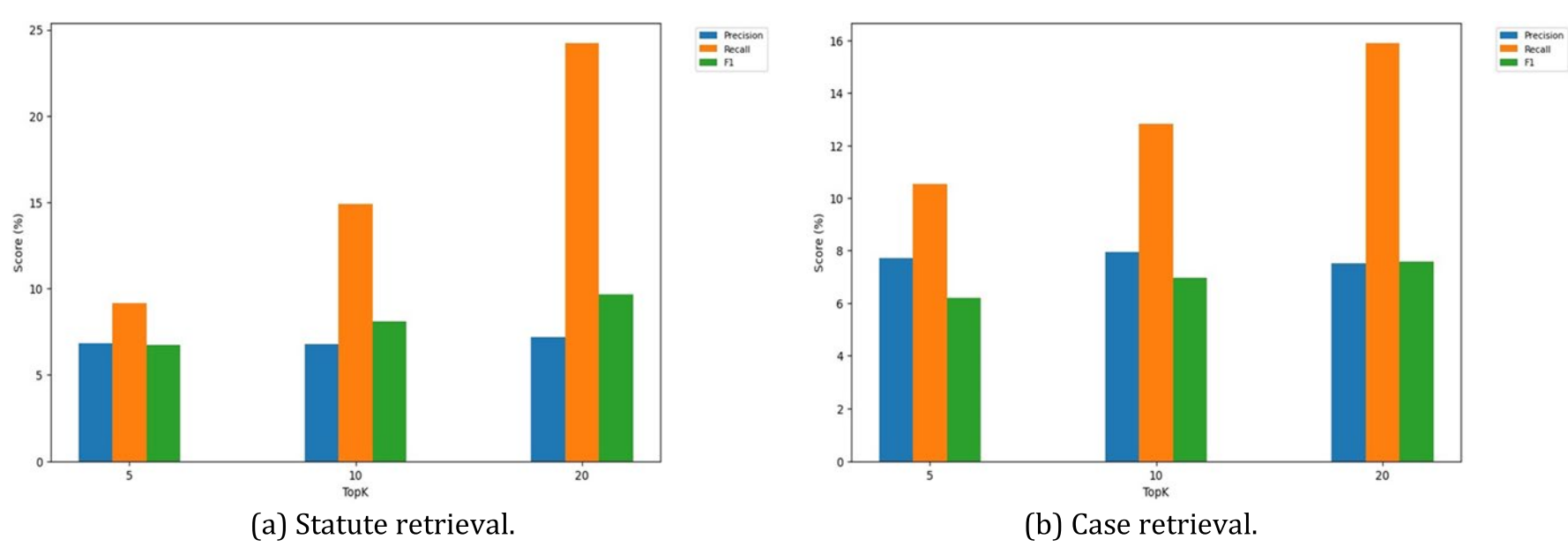

(a) Statute retrieval. (b) Case retrieval.

**Fig. 14** Retrieval results with varying topK using smooth TF-IDF

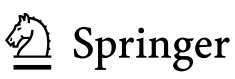

### 4.5 Application generation

We investigate the effectiveness of utilizing issues and rules for application generation. We reuse the same four LLMs in the previous stages and apply the prompt (Fig. 15 to generate legal analysis. The figure shows different components of the prompt. For different experimental settings, we sometimes remove the issues or self-evaluation prompt to compare the result.

#### 4.5.1 Evaluation details

Similar to issue generation, we apply GPT-3.5 turbo to compare the generated application sections with the annotated reference for each scenario. The possible outcomes of the evaluation include strongly agree (1), neutral (0), or disagree (−1). We also conduct a similar human evaluation to assess the quality of this automatic evaluation metric, and obtain a correlation of 0.86. Figure 16 displays the structure of the application evaluation prompt. The evaluation is based on the evaluation guidelines. The Fig. 17 shows the example output form the given LLM.

#### 4.5.2 Results

To provide a comprehensive analysis, we have introduced a baseline scenario without input issues and rules. Figure 18 shows the results when the prompts were incrementally added with issues and rules. Notably, all LLMs greatly benefit from the identified issues except for Gemini. GPT-3.5 turbo shows the most significant increase in performance, with an improvement of 18.9% from the self-evaluation prompt to the self-evaluation prompt with issues and rules. Generated applications are effective in improving the quality of conclusions, detailed in 4.6. We can also observe the improvement without adding the rule and issues.

Even though we provided a set of correct rules to LLMs, they still produced substantial errors in applying those rules. One of the common mistakes was associating wrong rules with facts in a reasoning step. For instance, a reasoning step generated by GPT-3.5 turbo states that “According to Malaysia Contract Law, Section 9(1), an advertisement is generally considered an invitation to treat, not an offer.” Although the statement “an advertisement is generally considered an invitation to treat” is correct, it has nothing to do with Section 9(1), even though the content of the rules are provided as a part of model inputs. The other common errors were caused by misunderstanding of rules, reasoning errors, or hallucinations such that logically implausible reasoning steps are generated.

For instance, in the reasoning step “*(3) IF Niko’s call to reserve that vinyl is an offer THEN Vanessa’s reply to reserve the vinyl for him until Wednesday 8 pm is*

Prompt for Application Generation

Instruction

Given a legal scenario and main questions, answer the main questions based on the decomposed question and rules and generate the analysis step by step in a point form. The response of the answer should follow the example of the sample analysis.

Input:

Scenario: [CONTENT OF THE SCENARIO]
Main legal question: [MAIN QUESTION PERTAINING TO THE SCENARIO]
Decompose question: [DECOMPOSE QUESTION LIST FROM GROUND TRUTH]
RULES : [SECTIONS AND CASES LIST]

Self-check instruction:

Based on the main question and generated analysis, select the most relevant one to the given scenario.

Sample analysis:

(1) According to Eckhardt Marine GmbH v Sheriff, High Court of Malaya, Seremban (2001)4 MLJ 49[10] THEN Alan's advertisement to sell the book for $1000 is an invitation to treat

(2) IF Alan's advertisement to sell the book for $1000 is an invitation to treat THEN Cate's reply to buy the book for $1000 is an offer. Section 4.1 In your response, please use a python list for the analysis.

Instruction

In your response, please use a python list for the analysis.

**Fig. 15** The prompt template for application generation with the most comprehensive inputs

Evaluation Prompt for Application Generation

Evaluation Criteria:

Do you agree that the analysis is a thorough examination of the facts and application of the law to the facts, consideration of counterarguments or alternative interpretations, and transparency regarding any assumptions made during the analysis?

Scoring Guidelines

1: Strongly agree - The document's analysis is logically structured and coherent, comprehensive, and makes reasonable assumptions. It addresses counterarguments and alternative interpretations.

0: Neutral - The analysis is logically structured but may lack comprehensiveness or may have some minor logical flaws according to the document.

-1: Disagree - The analysis lacks clear logical structure

**Fig. 16** The evaluation prompt for application generation

*an acceptance, according to* Sect. 4.1.", both the rule and the derived conclusion is wrong. The correct reasoning should be "*According to Section 7a, an acceptance must be absolute and unqualified. SINCE Vanessa's response is not absolute and unqualified THEN there is no acceptance.*".

## 4.6 Conclusion generation

We apply the same LLMs to generate conclusions, based on given scenarios, main issues, and reference application sections. The details of the prompt can be found in According to {Malaysia Contract Law, Section 9(1)}, an advertisement is generally considered an invitation to treat, not an offer.

**IF** the advertisement placed by Vanessa was an invitation to treat **THEN** Niko's call to reserve the vinyl is an offer.

**IF** Niko's call to reserve that vinyl is an offer **THEN** Vanessa's reply to reserve the vinyl for him until Wednesday 8pm is an acceptance. {Section 4.1} the conclusion prompt, we include the scenario, main questions, and the ground truth of the analysis, Fig. 19 in shows the details of the structure. We then ask the LLMs to produce the most reasonable conclusion. The constraint for the conclusion is that it must be in one sentence, answering the main question based on the analysis. According to Kang's work (Kang et al. 2023), providing intermediate steps helps improve the conclusion result. Therefore, we add the ground truth of the reasoning step to obtain the conclusion.

**Fig. 17** The llm output example from the given prompt

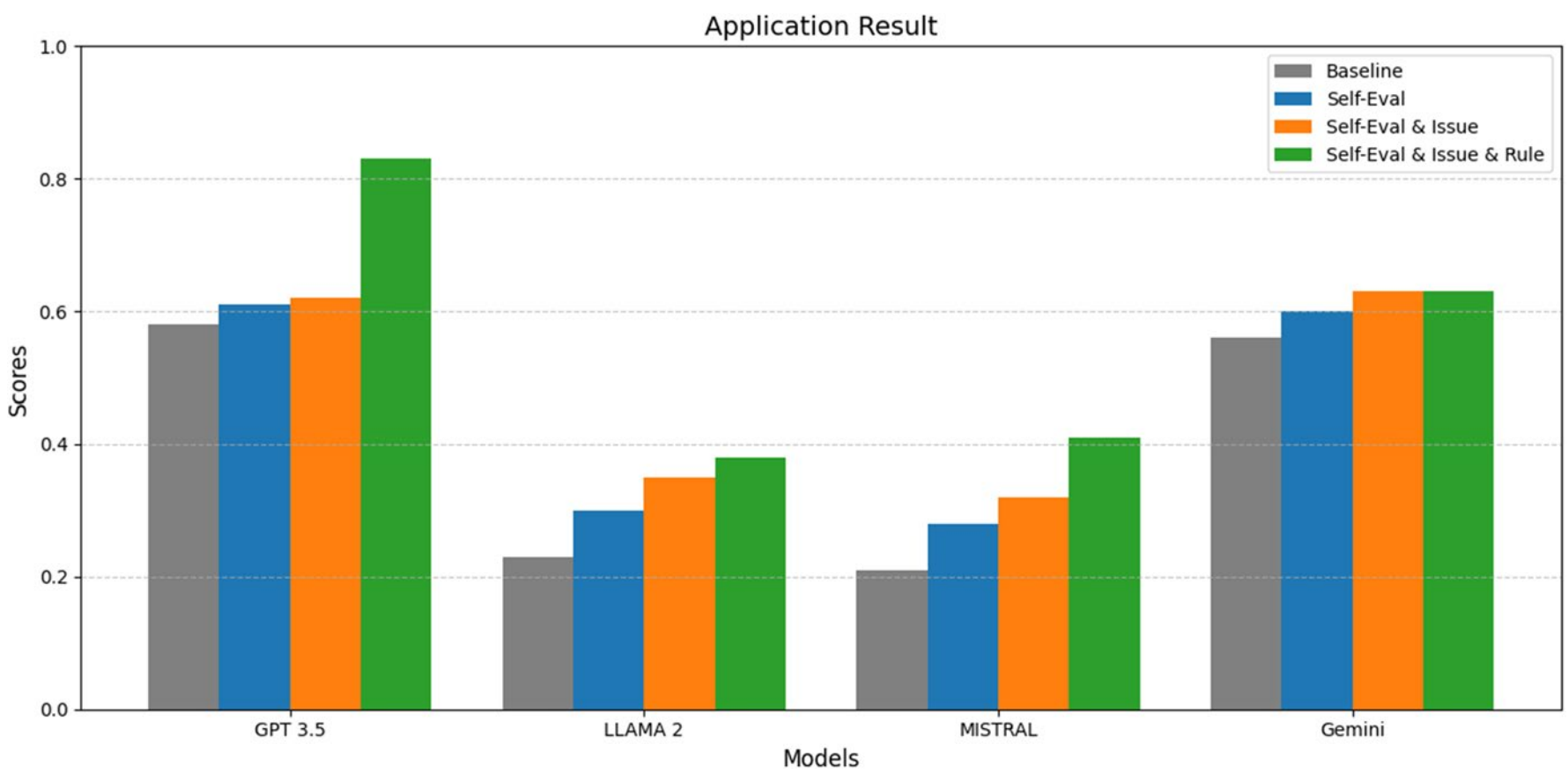

**Fig. 18** Application result

### 4.6.1 Conclusion Generation Evaluation

Figure 20 displays the structure of the conclusion evaluation prompt. The evaluation is based on the evaluation guidelines.

Prompt for conclusion

**Instruction:**

Given a legal scenario occurring in Malaysia and the main questions and analysis, please give a conclusion for the main question.

**Input:**

Scenario: [CONTENT OF THE SCENARIO]
Main legal question: [MAIN QUESTION PERTAINING TO THE SCENARIO]
Decompose question: [DECOMPOSE QUESTION LIST FROM GROUND TRUTH]
Application: [APPLICATION OF THE GIVEN SCENARIO]

Instruction:

In your response, please use a python list for the conclusion, the conclusion should be in one sentence.

**Fig. 19** The prompt template for conclusion generation with the most comprehensive inputs. Bold text denotes structural components that are omitted in the actual system prompt

| Evaluation Prompt for Conclusion Generation |
| --- |
| **Evaluation Criteria:** |
| Do you agree that the conclusion is Clear and concise? Alignment of the conclusion with the analysis and legal concepts. |
| Scoring Guidelines: |
| 1: Strongly agree - The document's conclusion is clear, concise, and aligns well with the analysis and legal concepts, addressing any unresolved issues or uncertainties.<br>0: Neutral - The conclusion is present but may need more clarity, conciseness, or better alignment with the analysis as per the document.<br>-1: Disagree - According to the document, the conclusion lacks clarity, conciseness, or alignment with the analysis. |

**Fig. 20** The evaluation prompt template for conclusion generation

### 4.6.2 Results

Figure 21 illustrates the comparative performance of four leading AI models (GPT-3.5, LLAMA 2, MISTRAL, and Gemini) under three evaluation conditions: baseline performance (blue), self-evaluation alone (orange) and self-evaluation with application (green). The baseline measurements represent each model's fundamental performance without additional evaluation methods, providing critical context for understanding relative improvements. The results demonstrate a consistent pattern across all models, with performance scores significantly improving when application is integrated with self-evaluation. GPT-3.5 exhibits the highest overall performance in both scenarios, achieving a base score of 0.32 with self-evaluation and rising dramatically to 0.58 when application is added—representing an 81% improvement. Similarly, Gemini shows the second-highest scores, with performance increasing from 0.18 to 0.41. LLAMA 2 and MISTRAL follow with more modest but still substantial improvements of 33% and 47%, respectively.

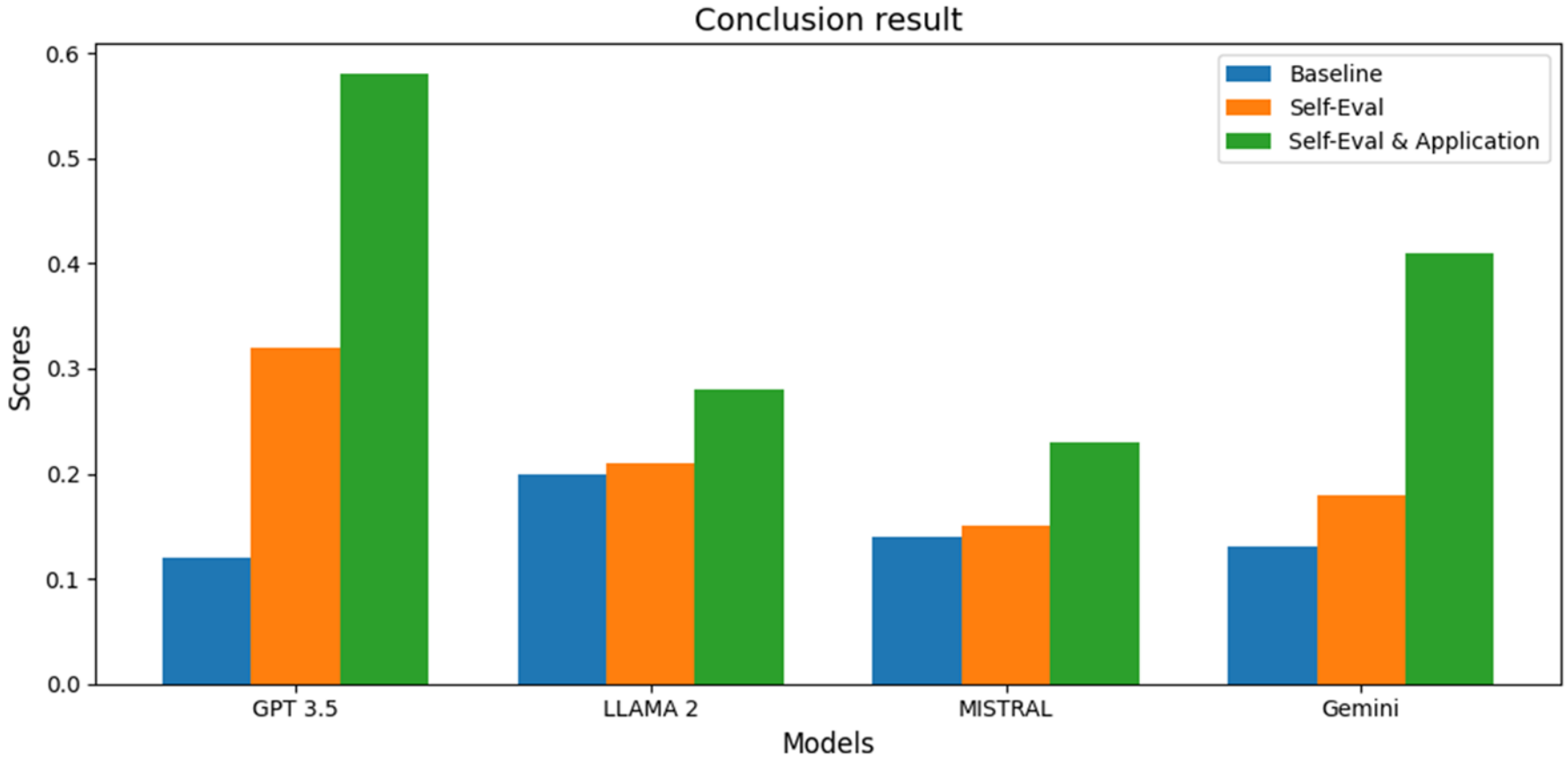

**Fig. 21** Result of conclusion

These findings strongly support the hypothesis that incorporating application components into self-evaluation processes substantially enhances the accuracy of model outputs. This alignment with previous research by Kang et al. (2023) suggests a robust pattern where contextual application of knowledge consistently improves model performance across different architectural designs.

## 5 Conclusion

We introduce LegalSemi, which consists of 54 scenarios annotated with IRAC analysis in Malaysian Contract Law, and a SKE for legal knowledge extracted from a law textbook and legislation. The SKE covers legal concepts, legal rules, interpretations in lay language and their relations. Legal concepts from the SKE are particularly useful for improving the quality of issue generation, which in turn significantly enhance legal analysis in Application across all four evaluated LLMs. By leveraging the SKE, we achieve a significant improvement in rule retrieval, with an increase of 17.2% in F1 score at top-5 by using enriched representations with illustrations and examples.

## Appendix 1

### Annotation guidelines

*Project Overview*

Develop a machine learning system for in-depth analysis of legal scenarios, specifically focusing on Contract Law utilising the IRAC (Issue, Rule, Analysis, and Conclusion) methodology.

*Methodology:*

Apply Contract Law principles to annotate data using the IRAC framework.

- *Project Requirements*Contract Law Expertise: A comprehensive understanding of Contract Law, particularly in relation to contract formation, is essential. You need to have B+ and above for the related subject.
- Responsibility and Time Management: Commitment to assigned tasks and timely completion is crucial.

- Basic IT Knowledge: Familiarity with computer systems and basic IT concepts is preferred.
- Communication and Teamwork: Strong communication skills and ability to collaborate effectively within a team are important.
- Pass the pre-test before starting the real annotation work.
- *Data Annotation Outcomes*
- Publication: The annotated dataset will be used for benchmarking and may be published in a journal or presented at a conference.
- Further Research: The annotated data will serve as a resource for subsequent machine learning research.
- *Benefits*
- Research Assistant Experience: Opportunity to work as a Data Annotator on a research project.
- Flexibility: Remote work with flexible hours.
- Compensation: RM 30 per hour.
- *Annotation Tasks*
- Evaluation of Legal Scenarios: Analyses and evaluate legal scenarios as per the IRAC framework.
- IRAC Analysis for Contract Formation: Apply IRAC methodology to analyse contract formation in provided scenarios.
- Decomposed Questions and Court Case References: Generate relevant decomposed questions for each IRAC segment and include related court cases with page numbers.

## Appendix 2

### Annotation tool

To facilitate this intricate annotation process, we developed an online data annotation platform, grounded in the principles of IRAC methodology. It is designed for universal accessibility, requiring only an internet connection. It features a 'Review' function, allowing annotators to refine and adjust their inputs as necessary. Data output is organized into a structured.json and./txt format, significantly enhancing efficiency and streamlining the data processing workflow for subsequent analysis. Figure 22shows an example of the annotation.

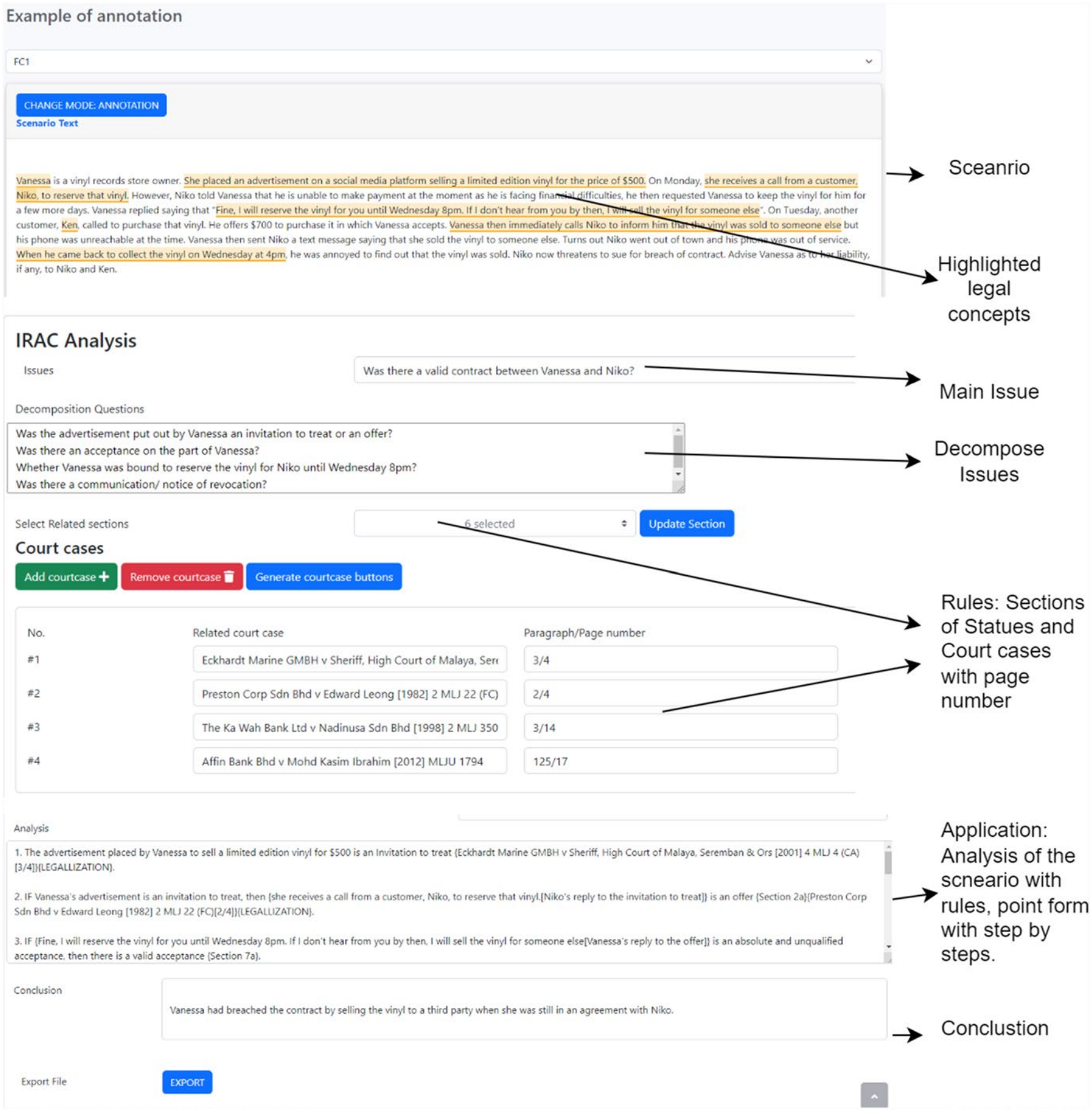

**Fig. 22** The web-based annotation tool developed to enable the legal scenario analysis

## Appendix 3

### Textbook details

Figure 23 showcases various sections and subsections that illustrate the organization of legal knowledge within the textbook. The index of the book's structured format, including headings, subheadings, and bullet points, mirrors the hierarchical nature of legal documents, making it conducive for rule-based knowledge extraction. At the end of the index is a link to the paragraph on that legal concept.

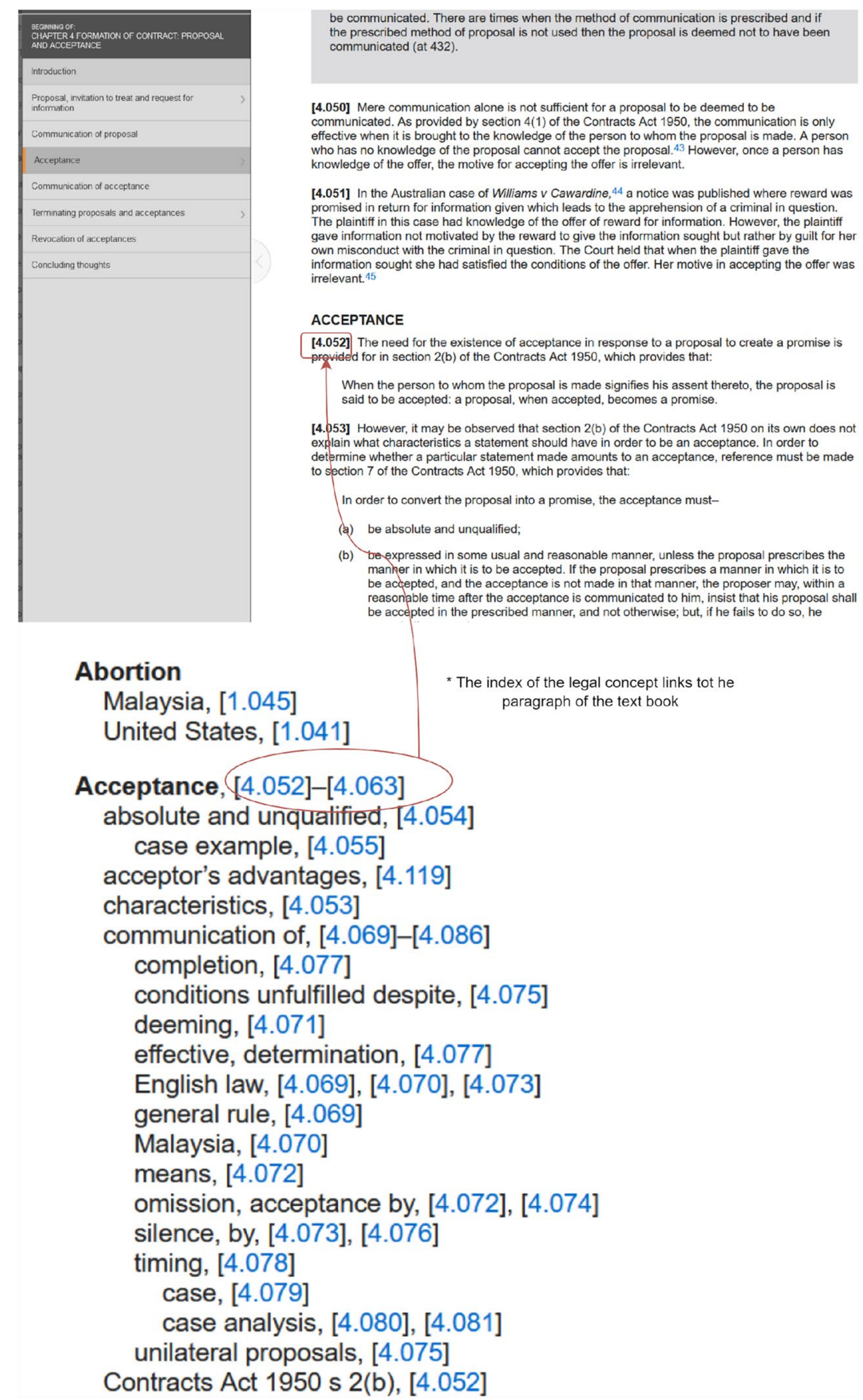

BEGINNING OF:
CHAPTER 4 FORMATION OF CONTRACT: PROPOSAL AND ACCEPTANCE

Introduction
Proposal, invitation to treat and request for information
Communication of proposal
Acceptance
Communication of acceptance
Terminating proposals and acceptances
Revocation of acceptances
Concluding thoughts

be communicated. There are times when the method of communication is prescribed and if the prescribed method of proposal is not used then the proposal is deemed not to have been communicated (at 432).

**[4.050]** Mere communication alone is not sufficient for a proposal to be deemed to be communicated. As provided by section 4(1) of the Contracts Act 1950, the communication is only effective when it is brought to the knowledge of the person to whom the proposal is made. A person who has no knowledge of the proposal cannot accept the proposal.[43] However, once a person has knowledge of the offer, the motive for accepting the offer is irrelevant.

**[4.051]** In the Australian case of *Williams v Cawardine*,[44] a notice was published where reward was promised in return for information given which leads to the apprehension of a criminal in question. The plaintiff in this case had knowledge of the offer of reward for information. However, the plaintiff gave information not motivated by the reward to give the information sought but rather by guilt for her own misconduct with the criminal in question. The Court held that when the plaintiff gave the information sought she had satisfied the conditions of the offer. Her motive in accepting the offer was irrelevant.[45]

**ACCEPTANCE**

**[4.052]** The need for the existence of acceptance in response to a proposal to create a promise is provided for in section 2(b) of the Contracts Act 1950, which provides that:

> When the person to whom the proposal is made signifies his assent thereto, the proposal is said to be accepted: a proposal, when accepted, becomes a promise.

**[4.053]** However, it may be observed that section 2(b) of the Contracts Act 1950 on its own does not explain what characteristics a statement should have in order to be an acceptance. In order to determine whether a particular statement made amounts to an acceptance, reference must be made to section 7 of the Contracts Act 1950, which provides that:

> In order to convert the proposal into a promise, the acceptance must–
>
> (a) be absolute and unqualified;
>
> (b) be expressed in some usual and reasonable manner, unless the proposal prescribes the manner in which it is to be accepted. If the proposal prescribes a manner in which it is to be accepted, and the acceptance is not made in that manner, the proposer may, within a reasonable time after the acceptance is communicated to him, insist that his proposal shall be accepted in the prescribed manner, and not otherwise; but, if he fails to do so, he

**Fig. 23** The screenshot of the structure of the textbook

## Appendix 4

### Human evaluation

Three human evaluators participated in the evaluation session. We selected 10 scenarios and two models (Gemini and GPT-3.5 turbo) with all the experiment settings for them to evaluate. We selected GPT-3.5 turbo and Gemini since they have the best performance from the auto evaluation result. The participants attended a briefing meeting to discuss and ascertain their understanding of the marking rubric. After the briefing, they independently evaluated the scenarios. The third evaluator, who is more experienced, served as the final decision-maker in cases where the first two annotators disagreed , ensuring the reliability and accuracy of the final results. This method follows the steps for identifying issues, which involves decomposing questions for this experiment. The remainder of the evaluation focuses on the application of these guidelines.

### Human evaluation guidelines

These guidelines are based on the marking rubric and evaluation criteria for contract law (Gerhardt 2008). They determine how and why the conclusion is made. The answer needs to demonstrate the facts, which include the description of the circumstances, the main questions, and the issues (decomposed questions) raised by the question. It should also cover the rules and principles related to the issues, as well as the conclusions drawn. The proper approach is similar to that of the court's judgments. The grade in Fig. 24 shows is a helpful step-by-step process to evaluate the legal reasoning process.

Decompose Questions

Fail

This response minimally identifies the relevant issues as outlined in the ground truth. It shows a superficial awareness of the primary topics required for discussion but lacks depth and detail in all areas. The response fails to differentiate between closely related ideas or apply the concepts accurately.

Example: It mentions the concept of an offer in a contract scenario but does not differentiate between an offer and an invitation to treat, providing vague and inaccurate references to the principles involved. Additionally, it does not adequately address the relevance of Questions 5 and 6 to Amy's capacity to enter into a contract.

Pass

This response correctly identifies most of the issues and demonstrates reasonable accuracy in articulating the relevant legal principles. It offers a clearer and more comprehensive understanding of the required legal knowledge compared to the ground truth, though it might not explore all nuances or less obvious aspects.

Example: It identifies and explains both the offer and acceptance in a sale of goods contract, but the analysis of consideration is brief and lacks a discussion on the adequacy of consideration or exceptions like past consideration.

Application

Credit

This response successfully identifies all relevant issues and states the law with reasonable accuracy, resolving the issues with some effectiveness. Its analysis is well-integrated and appropriately linked to the stated legal principles, showing good application of these principles as reflected in the ground truth.

Example: A student comprehensively identifies offer, acceptance, consideration, and capacity to contract, with accurate references to Malaysian legal principles. The analysis correctly applies these principles to the facts but may not fully address complex issues like implied terms or conditions.

Distinction

This response accurately identifies all issues, provides substantial accuracy in stating the law, and resolves issues with clear, logical reasoning. It demonstrates a thorough understanding and application of the law, with detailed references to supporting cases and statutes, aligning closely with the ground truth analysis.

Example: It provides a detailed discussion of offer, acceptance, consideration, capacity, and intention to create legal relations, with substantial accuracy. The response includes detailed examples and references to key Malaysian cases and statutory provisions, offering a nuanced analysis of each element.

High Distinction

This response is near-perfect, excelling in identifying issues, stating the law with high accuracy, and resolving issues effectively. Its analysis is exceptionally thorough, well-reasoned, and demonstrates an outstanding understanding of the legal framework, often surpassing the ground truth in its depth and logical evaluation.

Example: It not only meticulously addresses all elements of contract formation with precise legal references and applications but also critiques ambiguities in the current law, suggests areas for reform based on comparative analysis with other legal systems, and discusses the implications of recent Malaysian case law on future contract formations.

Conclusion

Entailment

The conclusion logically follows from the given premises or statements.

Controdiction

The conclusion directly opposes or negates the given premises or statements.

Example: There is a valid contract between Vanessa and Niko.
Result: Controdiction

**Fig. 24** Human evaluation guidelines

**Acknowledgements** Part of this research is supported by the Monash PhD Global Mobility Program, which enabled collaboration with esteemed experts from the Faculty of Information Technology, Monash University Australia. This project is also partly supported by the Malaysian Fundamental Research Grant Scheme (FRGS) FRGS/1/2022/ICT02/MUSM/02/2.

**Funding** Open Access funding enabled and organized by CAUL and its Member Institutions.

## Declarations

**Ethical** Our research practices strictly adhere to the principles of Monash Ethics Integrity and have been approved by the Monash Ethics Office under project number 44946. This approval underscores our commitment to conducting research in compliance with the highest ethical standards.
The development and evaluation of LegalSemi were carried out with a strong focus on ethical considerations, particularly regarding the involvement of human annotators. We recognize the significant effort required for generating human-annotated data to train conditional independence classifiers in our method. To ensure this process is ethically sound, we have implemented measures to honor the contributions of annotators, including fair and transparent compensation and respect for their time. Additionally, the primary goal of LegalSemi is to establish an IRAC methodology-based benchmark that prioritizes ethical integrity. It is explicitly designed to avoid generating or disseminating any information that could harm individuals or compromise privacy. This conscientious approach reflects our dedication to fostering a research environment that upholds both ethical responsibility and societal benefit.

## Authors and Affiliations

**Xiaoxi Kang[1] · Lizhen Qu[2] · Lay-Ki Soon[1] · Zhuang Li[4] · Adnan Trakic[3]**

✉ Xiaoxi Kang
xiaoxi.kang@monash.edu

✉ Lizhen Qu
lizhen.qu@monash.edu

Lay-Ki Soon
soon.layKi@monash.edu

Zhuang Li
zhuang.li@rmit.edu.au

Adnan Trakic
adnan.trakic@monash.edu

1 School of Information Technology, Monash University Malaysia, Jalan Lagoon Selatan, Bandar Sunway, 47500 Subang Jaya, Selangor, Malaysia

2 Department of Data Science & AI, Monash University Australia, Wellington Rd, Clayton, VIC 3800, Australia

3 School of Business, Monash University Malaysia, Jalan Lagoon Selatan, Bandar Sunway, 47500 Subang Jaya, Selangor, Malaysia

4 School of Computing Technologies, RMIT University, La Trobe Street, Melbourne, VIC 3000, Australia